\documentclass{article}

\usepackage[preprint]{neurips_2025}
\usepackage{wrapfig}
\usepackage{microtype}
\usepackage{graphicx}
\usepackage{amsfonts}
\usepackage{subfigure}
\usepackage{booktabs} % for professional tables
\usepackage{afterpage}
\usepackage{algorithm,algpseudocode}

\usepackage{comment}
\usepackage{booktabs}
\usepackage{adjustbox}
\usepackage{multirow}
\usepackage{amsmath}
\usepackage{amssymb}
\usepackage{mathtools}
\usepackage{amsthm}
\usepackage{xcolor}
\theoremstyle{plain}
\newtheorem{theorem}{Theorem}%[section]

\newtheorem{lemma}{Lemma}%[section]

\theoremstyle{definition}

\theoremstyle{remark}

\usepackage[textsize=tiny]{todonotes}
\usepackage[utf8]{inputenc} % allow utf-8 input
\usepackage[T1]{fontenc}    % use 8-bit T1 fonts
\usepackage{hyperref}       % hyperlinks
\usepackage{url}            % simple URL typesetting
\usepackage{booktabs}       % professional-quality tables
\usepackage{amsfonts}       % blackboard math symbols
\usepackage{nicefrac}       % compact symbols for 1/2, etc.
\usepackage{microtype}      % microtypography
\usepackage{xcolor}         % colors
\usepackage{multirow}
\usepackage{makecell}

\title{When can isotropy help adapt LLMs' next word prediction to numerical domains?}
\author{
Rashed Shelim\\
Department of Electrical and Computer Engineering \\
\& Department of Computer Science\\
Virginia Tech\\
\texttt{rasheds@vt.edu} \\
\And
Shengzhe Xu  \\
Department of Computer Science\\
Virginia Tech\\
\texttt{shengzx@vt.edu} \\
\And
Walid Saad  \\
Department of Electrical and Computer Engineering\\
Virginia Tech\\
\texttt{walids@vt.edu} \\
\AND
Naren Ramakrishnan \\
Department of Computer Science\\
Virginia Tech\\
\texttt{naren@cs.vt.edu} 
}

\begin{document}

\maketitle
\vspace{-0.2in}

\begin{abstract}
\vspace{-0.05in}
 Vector representations of contextual embeddings learned by pre-trained large language models (LLMs) are effective in various downstream tasks in \emph{numerical domains} such as time series forecasting.  Despite their significant benefits, the tendency of LLMs to hallucinate in such domains can have severe consequences in applications such as energy, nature, finance, healthcare, retail and transportation, among others. To guarantee prediction reliability and accuracy in numerical domains, it is necessary to open the black box behind the LLM and provide performance guarantees through explanation. However, there is little theoretical understanding of when pre-trained language models help solve numerical downstream tasks. This paper seeks to bridge this gap by understanding when the next-word prediction capability of LLMs can be adapted to numerical domains through a novel analysis based on the concept of isotropy in the contextual embedding space. Specifically, a log-linear model for LLMs is considered in which numerical data can be predicted from its context through a network with softmax in the output layer of LLMs (i.e., language model head in self-attention). For this model, it is demonstrated that, in order to achieve state-of-the-art performance in numerical domains, the hidden representations of the LLM embeddings must possess a structure that accounts for the shift-invariance of the softmax function. By formulating a gradient structure of self-attention in pre-trained models, it is shown how the isotropic property of LLM embeddings in contextual embedding space preserves the underlying structure of representations, thereby resolving the shift-invariance problem and providing a performance guarantee. Experiments show that different characteristics of numerical data and model architectures have different impacts on isotropy, and this variability directly affects the performances.
 %Experiments across $22$ different numerical datasets and $5$ different language models show that different characteristics of numerical data and model architectures could have different impacts on the isotropy measures, and this variability directly affects the time series forecasting performances.
\end{abstract}
\vspace{-0.19in}
\section{Introduction}
\vspace{-0.13in}
Large language models (LLMs) have been proven to effective in adapting to various downstream tasks in numerical domains, such as finance~\cite{garza2023timegpt,yu2023temporal}, energy~\cite{gao2024inducing}, climate science~\cite{jin2023time}, healthcare~\cite{wang2024large}, transportation signals~\cite{xu2024large}, synthetic tabular generation~\cite{dinh2022lift,borisov2023language,xu2024large}, among others. Inspired by the success of pre-trained LLMs, several methods have been developed recently in~\cite{gruver2024, dooley2023, nie2023a,  rasul2024, woo2024, jin2023time, ansari2024chronos} by adapting LLM to numerical domains that deal with time series forecasting. For many of these numerical downstream tasks, training a linear classifier on top of the hidden-layer representations generated by the pre-trained LLMs has been shown to achieve near state-of-the-art performance \cite{jin2023time,ansari2024chronos}. However, the existing models in~\cite{gruver2024, dooley2023, nie2023a,  rasul2024, woo2024, jin2023time, ansari2024chronos} are treated as a `black box' where numerical forecasts are controlled by complex nonlinear interactions between many parameters. This makes it difficult to understand how models arrive at their predictions and makes it challenging for users to trust the model outputs.  

When applied to critical numerical domain use cases, the tendency of LLMs to hallucinate can have serious and detrimental consequences. For example, prediction errors in fraud detection in finance can lead to huge financial losses and errors in protection onset of sepsis or cardiac arrest in healthcare can result in patient deaths. Thus, to guarantee prediction reliability and accuracy in numerical domains, it is necessary to understand the inner working of the so-called black box and provide performance guarantees through explanation. Although recent empirical studies \cite{jin2023time, nie2023a, liu2024llm4cp} demonstrate the benefits of vector representations of embedding learned by LLMs in various numerical downstream tasks, there is little theoretical understanding of their empirical success. 
Thus, a fundamental question arises: \textit{``When (or how) can the next-word prediction capability of LLMs be effectively adapted to numerical domains?"}

The main contribution of this paper is a novel approach for answering this question by exploiting the isotropic property of LLM hidden representations in the contextual embedding space. %the lens of \emph{isotropy}. 
\emph{Isotropy} refers to the geometric property wherein vector representations in the embedding space are uniformly distributed in all directions, a characteristic critical for maintaining the expressiveness of the embedding space \cite{arora-etal-2016-latent, mu2018allbutthetop}. 
To achieve state-of-the-art performance in numerical domains, we show that the hidden representations of LLMs must exhibit \emph{a structured form} in contextual embedding space that accounts for the shift-invariance of the softmax function (i.e., the softmax output remains unchanged when all logits are shifted by a constant). Without such structure, the model can shift the logits while keeping the training loss unchanged, thereby leaving the logits ineffective for numerical downstream tasks. By formulating a gradient structure of self-attention in pre-trained models, we show how the isotropic property of LLM embeddings in the contextual embedding space preserves the underlying structure of representations, thereby resolving the shift-invariance problem of the softmax function. In a nutshell, our key contributions include:
\vspace{-0.1in}
\begin{itemize}
    \item We consider a log-linear model for LLMs and demonstrate theoretically why hidden representations must exhibit structure to address the shift-invariance problem of the softmax function. 
    \item We take a deeper look into the hidden representations of pre-trained models and show how isotropy preserves the structural integrity of representations. In particular, we derive an upper bound for the Jacobian matrix which collects all first-order partial derivatives of self-attention with respect to the input pattern and show that the $m$ largest eigenvectors of the LLM hidden representations minimize the gradient norm of self-attention. Then, by projecting the representations into lower dimensions using these $m$ largest eigenvectors, we find the isotropy within the clusters in the contextual embedding space. 
    \item Finally, we provide a comprehensive evaluation across $12$ real and $10$ synthetic time series datasets over $5$ different LLMs. Our experiments demonstrate that the isotropy of LLM hidden representations varies significantly based on the input data characteristics (i.e., domain, context length and noise level) and model design choices (i.e., tokenization techniques and architecture), which in turn strongly influences forecasting performance in numerical domains.
\end{itemize}
\vspace{-0.15in}
\section{Problem Setup in Numerical Domains}\label{model}
\vspace{-0.12in}
\textbf{Time Series Tokens and Similarity Measure.} 
Similar to next-word prediction by LLMs, the next-value prediction in the numerical domain can be modeled by \textit{time series forecasting} techniques which are widely adopted in the machine learning literature \cite{jin2023time,ansari2024chronos}.
Formally, given a time series  $\textbf{x}_{1:T+L}=[x_1,\ldots,x_T,\ldots,x_{T+L}]$,
\begin{wrapfigure}[15]{r}{0.35\textwidth}%\label{timesrs}
\centering
\includegraphics[width=.2\textwidth]{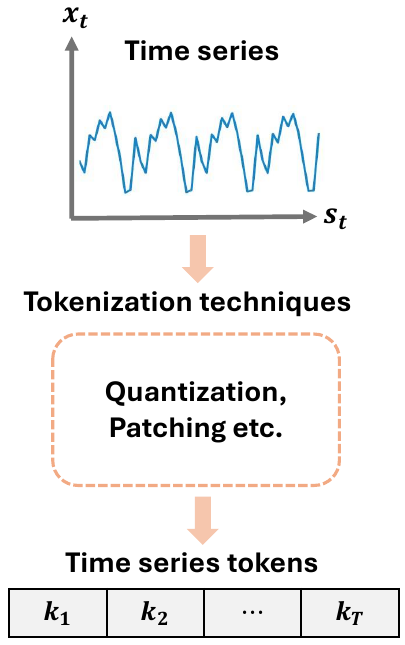}
%\vspace{-4pt}
\caption{Time series tokenization.}
\label{timesrs}
\end{wrapfigure}
where the first $T$ time instances give the historical context, the next $L$ time instances constitute the forecast region, and $x_t\in \mathbb{R}$ is the observation of each time instance, we are interested in predicting the joint distribution of the next $L$ time instances, $p(\textbf{x}_{T+1:T+L}|\textbf{x}_{1:T})$. Since, the pre-trained models operate on tokens from a finite vocabulary, using them for time series data requires mapping the observations to a finite set of tokens. 
Based on different numerical applications and LLM architectures, various tokenization techniques, e.g., quantization and scaling \cite{ansari2024chronos,rasul2024}, patching \cite{woo2024,jin2023time, nie2023a}, and adaptation of language model tokenizer in numerical domians~\cite{gruver2024, dooley2023}, can be applied to tokenize the time series and create a time series vocabulary $\mathcal{V}$ of $N$ time series tokens, i.e., $|\mathcal{V}|=N$, as shown in Figure~\ref{timesrs}. Then, the realization of the next $L$ time instances can be obtained by autoregressively sampling from the predicted distribution $p(k_{T+l+1}\,|\, \bold{k}_{1:T+l})$, for $l\in\{1,\ldots,L\}$, where $\bold{k}_{1:T+l}$ is the tokenized time series and 
%\textbf{Metric.} The metric we use for measuring isotropy in LLM representations in the contextual embedding space is cosine similarity metric. Let 
$k_i$ be a time series token in time series vocabulary $|\mathcal{V}|$. %We call the time series token a \textit{token} for ease of reading. 

Let %each token $i$ in $\mathcal{V}$ is represented by $k_i$, and 
$\tilde{\bold{\Psi}}(k_i)=\{\bold{\psi}_1(k_i),\bold{\psi}_2(k_i),\ldots\}$ be the set of all LLM contextual embedding instances of time series token $k_i$. Here, different contexts in the time series sequences yield different LLM embeddings of $k_i$. 
By constructing $\sum_{k}|\tilde{\bold{\Psi}}(k)|=|\mathcal{V}|$, we define the inter-token cosine similarity as: 
%\vspace{-0.4in}
\begin{align}%\label{cos_sim}
\zeta_{\textrm{cos}} \triangleq  \mathbb{E}_{i \neq j} [\cos (\bold{\psi}(k_i), \bold{\psi}(k_j))],\label{cosine_nocluster}
\vspace{-0.9in}
\end{align}
where $\bold{\psi}(k_i)$ and $\bold{\psi}(k_j)$ are random samples from $\tilde{\bold{\Psi}}(k_i)$. %, and the same for $\bold{\psi}(k_j) \in \bold{\psi}(k_j)$. 
The expectation is taken over all pairs of different tokens. The inter-token cosine similarity metric describes the similarity between different tokens based on the contexts. For notational simplicity, we express $T+l$ as $T_{l}$ and $T+l+1$ as $T_{l+1}$ hereinafter.

\textbf{Model.} We consider a general pre-trained model for numerical data and open the black box of the pre-trained model. Specifically, we assume that the observation probability of $k_{T_{l+1}}$ given $\bold{k}_{1:T_l}$ satisfies the log-linear model as in \cite{arora-etal-2016-latent}
\vspace{-0.05in}
\begin{align}
    p^*(k_{T_{l+1}} \!= \!i \,\,|\,\, \bold{k}_{1:T_l}) \propto \exp(\langle \bold{\bold{\psi}}^*(\bold{k}_{1:T_l}), \bold{\psi}^*(k_i)\rangle), \label{loglin}
\end{align}
%\vspace{-0.08in}
where $\bold{\psi}^*(k_i)\in \mathbb{R}^D$ is a vector that only depends on the time series token $k_{i\in\mathcal{V}}$, and $\bold{\bold{\psi}}^*(\bold{k}_{1:T_l})$ is a function that encodes the tokenized time series sequence $\bold{k}_{1:T_l}$ into a vector in $\mathbb{R}^D$. The log-linear modeling aligns with the commonly used LLMs networks whose last layer is typically a softmax layer. Moreover, we do not consider any prior distribution for input, which makes our model more general than previous latent models  \cite{arora-etal-2016-latent, NEURIPS2021_86b3e165}. 

To define the numerical downstream task, let $z^*_i(k,l)\!\!:=\!\!\langle \bold{\bold{\psi}}^*(\bold{k}_{1:T_l}), \bold{\psi}^*(k_i)\rangle$ be the $i$-th logit of the ground-truth model, and assume that the numerical downstream tasks are defined by a function of the logits, i.e., $f^*(\bold{z^*})$. Also let $Z^*(k,l)\! \!= \!\!\sum_{i=1}^{N} \exp(z^*_i(k,l))\!=\! \sum_{i=1}^{|\mathcal{V}|} \exp(\langle \bold{\bold{\psi}}^*(\bold{k}_{1:T_l}), \bold{\psi}^*(k_i)\rangle)$ be the partition function \cite{arora-etal-2016-latent}, i.e., normalization factor. In LLMs, the partition function is often used to normalize the output probabilities of the model, ensuring that they sum to $1$. %For our case, we use the partition function $Z(k,l)$ to normalize the ground-truth model in \eqref{loglin} and use this normalized model in training. 
Then, the normalized ground-truth model $\forall i\!\in\!\mathcal{V}$ is given by
\vspace{-0.06in}
\begin{align}
p(k_{T_{l+1}}\!\!= i\,\, | \,\,\bold{k}_{1:T_l}) &=\!\frac{\exp(\langle \bold{\bold{\psi}}^*(\bold{k}_{1:T_l}), \bold{\psi}^*(k_i)\rangle)}{\sum_{i=1}^{|\mathcal{V}|} \exp(\langle \bold{\bold{\psi}}^*(\bold{k}_{1:T_l}), \bold{\psi}^*(k_i)\rangle)}\nonumber= \frac{\exp(z^*_i(k,l))}{Z^*(k,l)}.
\end{align}
%for $\forall j\!\in\!\mathcal{V}.$
%We use the normalized model in \eqref{norm_loglin} in training.
\vspace{-0.15in}

%\textbf{Student Model.} 
Since we do not know the ground-truth model in reality, we do not have access to the ground-truth model components $\bold{\psi}^*(k_i)$ and $\bold{\bold{\psi}}^*(\bold{k}_{1:T_l})$. Instead, we only have access to the student model $\bold{\psi}(k_i)$ and $\bold{\bold{\psi}}(\bold{k}_{1:T_l})$ that aims to achieve low pre-training loss. We can define the student logits as $\bold{z}(k,l)\!\!:=\!\!\{\langle \bold{\bold{\psi}}(\bold{k}_{1:T_l}), \bold{\psi}(k_i)\rangle\}_{i=1}^{|\mathcal{V}|}$. Intuitively, $\bold{z}$ are the contextualized representations learned by the student-model during pre-training. Then, the solution of the downstream task is to learn a function $f(k,l)$. %Hence, similar to the ground-truth model, we use $z_i(k,l)\!\!:=\!\!\langle \bold{\bold{\psi}}(\bold{k}_{1:T_l}), \bold{\psi}(k_i)\rangle$ to denote the $i$-th logit and $Z(k,l)\! \!= \!\!\sum_{i=1}^{\mathcal{|V|}} \exp(z_i(k,l))$ to denote the partition function. For our case, we use the partition function $Z(k,l)$ to normalize the student model and use this normalized model in training. 
Then, the output of the student model $\forall i\!\in\!\mathcal{V}$ can be defines as
\vspace{-0.06in}
\begin{align}
p\,(k_{T_{l+1}} = i\,\, | \,\,\bold{k}_{1:T_l}) \!&=\! \frac{\exp(\langle \bold{\bold{\psi}}(\bold{k}_{1:T_l}), \bold{\psi}(k_i)\rangle)}{Z(k,l)}.\label{norm_loglin}
\vspace{-1.5in}
\end{align}

\textbf{Loss Function.}
As typical in language models, we use the categorical distribution over the elements in the time series vocabulary $\mathcal{V}$ as the output distribution  $p(k_{T_{l+1}} | \bold{k}_{1:T_l})$, for $l\in\{1,\ldots,L\}$, where $\bold{k}_{1:T_l}$ is the tokenized time series. The student model is trained to minimize the cross entropy between the distribution of the tokenized ground truth label and the predicted distribution. The loss function for a single sequence of tokenized time series is given by \cite{ansari2024chronos,wu2023connecting}
\vspace{-0.08in}
\begin{align}
%\ell(v_{1:T_l})
\mathcal{L}&= \!-\! \sum_{l=1}^{L+1}\!\sum_{i=1}^{|\mathcal{V}|} %\textbf{1}_{(k_{\tau_{l+1}} \!= \!j)} 
p^*(k_{T_{l+1}} \!=\!i \!\,\,|\!\,\, \bold{k}_{1:T_l})\log p(k_{T_{l+1}} \!= \!j \,\,\!|\!\,\, \bold{k}_{1:T_l})\nonumber\\
&= \sum_{l=1}^{L+1}\!\sum_{i=1}^{|\mathcal{V}|} \!\mathcal{D}_{\textrm{KL}} ( p^*(k_{T_{l+1}} \!= \!i|\bold{k}_{1:T_l}) \| \,p(k_{T_{l+1}} \!= \!j| \bold{k}_{1:T_l}))+ \!H(p^*(k_{T_{l+1}}\!=\!j|\bold{k}_{1:T_l})),\label{KL}
\vspace{-5in}
\end{align}
where $p(k_{T_{l+1}} \!\!= \!i \,|\, \bold{k}_{1:T_l})$ is the categorical distribution predicted by the student model parametrized by $v_{1:T_l}$, $p^*(k_{T_{l+1}} \!\!= i | \bold{k}_{1:T_l})$ is the distribution of ground-truth model, $\mathcal{D}_{\textrm{KL}}$ is the KL divergence%, i.e., weighted log probability difference between the ground-truth and the student model
, and $H(p^*(k_{T_{l+1}} \!\!= \!i |\, \bold{k}_{1:T_l}))$ is the entropy of distribution $p^*(k_{T_{l+1}} \!= \!i |\bold{k}_{1:T_l})$ which is a constant. %Note that our model performs regression via classification \cite{TORGO1997275} through the categorical entropy loss in \eqref{KL}. 
We assume that student model achieves a small loss so that the KL-divergence term in \eqref{KL} is also small. 
%\vspace{-0.04in}

\textbf{Downstream Numerical Task.}
%The numerical downstream task that we are considering is \emph{regression via classification} (as described in the previous section). 
We consider a simple downstream task whose prediction on categorical distribution is linear in $\bold{\bold{\psi}}^*(\bold{k}_{1:T_l})$, that is, $f^*(k,l)=\langle \bold{\bold{\psi}}^*(\bold{k}_{1:T_l}), u^* \rangle= \sum_{i=1}^{|\mathcal{V}|}a^*_iz^*_i(k,l),$ where $u^*=\sum_{i=1}^{|\mathcal{V}|}a^*_i\bold{\psi}^*(k_i)\in\mathbb{R}^D$ and $a_j$ is the coefficient. This model is still not sufficient to provide a performance guarantee to generalize to downstream task in unseen scenarios. %This is due to the fact that, for the entries with small ground-truth probabilities, a large log probability difference does not results in a large KL divergence in the loss function in \eqref{KL}. 
However, the log probability difference is proportional to the difference in the value of the perfect model (i.e., ground-truth) $f^*(k,l)$. This allows the student model to alter the signs of $f^*(k,l)$ without resulting in a large KL divergence \cite{wu2023connecting}. %Therefore, additional conditions must be applied to reduce the sensitivity of numerical downstream tasks to small logits. 
Then, it is more reasonable to model the numerical downstream task as
\vspace{-0.084in}
\begin{align}
f^*(k,l)\nonumber&= \sum_{i=1}^{|\mathcal{V}|}a^*_i\sigma(z^*_i(k,l)-b^*_i) = \sum_{i=1}^{|\mathcal{V}|}a^*_i\sigma(\langle \bold{\bold{\psi}}^*(\bold{k}_{1:T_l}), \bold{\psi}^*(k_i)\rangle-b^*_i),\label{good_ml}
\vspace{-1.5in}
\end{align}
%\vspace{-1.5in}
where $\sigma$ is the ReLU function and $b^*_j$ denotes the threshold for the logits. The numerical downstream task only considers the logits that are above the threshold, and thus ignores all the entries with very small probabilities.

Despite the empirical success of language models in numerical domains, there remains a fundamental gap in understanding when and why these models generalize reliably to numeric downstream tasks such as time series forecasting across different numerical domains. A key challenge lies in the mismatch between training and inference behavior, i.e., good training performance does not always guarantee robust performance at inference time. To address this challenge, we propose a novel theoretical framework grounded in the isotropic property of the contextual embedding space. We show that the presence of strong isotropy in LLM hidden representations stabilizes the partition function, effectively resolving the softmax shift-invariance problem and leading to reliable inference performance. The next section formalizes this insight and provides theoretical justification for using isotropy as a key indicator of model reliability in numerical settings.

%The next section formalizes this insight and presents theoretical justification for using isotropy as a measure of structural  of both model reliability and robustness in numerical settings.
%Despite the empirical success of language models in numerical domains, there remains a fundamental gap in understanding when and why these models generalize effectively to numeric downstream tasks such as time series forecasting. A critical challenge lies in the fact that good training performance does not necessarily translate to reliable behavior at inference time. In this work, we address this challenge by proposing a novel theoretical framework based on the isotropic property of the contextual embedding space. We show that the existence of strong isotropy in LLM's hidden representations stabilizes the partition function, thereby resolving the softmax shift-invariance problem and enabling consistent performance. The next section formalizes this insight and provides theoretical justification for using isotropy as a key indicator of model reliability in numerical domains.
%\vspace{-0.15in}
\section {The Role of Isotropy in Adapting LLMs to Numerical Data}\label{iso}
\vspace{-0.1in}
As discussed in Section~\ref{model}, we consider LLM networks whose last layer is usually a softmax layer and the numerical downstream task
is determined by the function of the logits. The underlying relation between the logits and softmax function determines the performance of the numerical downstream tasks. However, the softmax function is shift-invariant, that is, the output of the softmax function remains unchanged when all logits are shifted by a constant. Since we do not have any control over the logit shift of the student model on unseen data, good performance during training does not necessarily provide any performance guarantee for the numerical downstream task on unseen scenarios. The lack of perfromance guarantee under uncontrolled logit shifts on unseed data can be formalized in the following theorem:  %and the proof is provided in Appendix~\ref{TH1}. %Theorem~\ref{th1} whose proof is provided in Appendix~\ref{Ap_th1}. 
%\vspace{-0.02in}
\begin{theorem}\label{th1}
Let the logits of the ground-truth model be bounded. Then for any $f^*(k,l)$, there exists a set of functions $\{\hat{z}_i(k,l)\}_{i=1}^{|\mathcal{V}|}$ such that for all $k$ and $T_{l+1}$, the predictive distribution of the student model $\hat{p}\,(k_{T_{l+1}}\, | \, \bold{k}_{1:T_l})$ matches that of ground-truth model $p^*(k_{T_{l+1}}\,|\,\bold{k}_{1:T_l})$ and $\hat{f}(k,l) = 0$. In other words, there exists a student model with the same pre-training loss as the ground-truth model, but its logits are ineffective for the numerical downstream tasks.
\end{theorem}
\vspace{-0.15in}
\begin{proof} 
The proof is provided in Appendix~\ref{TH1}.
\end{proof}
\vspace{-0.1in}
%\vspace{-0.08in}
Theorem~\ref{th1} shows that without any structure in the hidden representations of LLM embeddings, student model can shift the logits for any sample while keeping the pre-training loss unchanged and leaving logits ineffective for the numerical downstream tasks. Consequently, a theoretical guarantee for  numerical downstream task performance will require structure in the LLM representations learned by the pre-trained model. 

In this paper, we make an observation that to prevent the shift-invariance problem from influencing the performance of the numerical downstream tasks, it is necessary to keep the partition function stable. 
Let $\bold{\Psi} \!=\! (\bold{\psi}_1(k), \ldots, \bold{\psi}_{|\mathcal{V}|}(k))^\top \!\in\! \mathbb{R}^{|\mathcal{V}| \times D }$ 
be the hidden representations of input time series sequence. Then, the stability of the partition function can be assessed through the isotropy in the contextual embedding space \cite{arora-etal-2016-latent,mu2018allbutthetop} as follows 
\vspace{-0.03in}
\begin{equation}
    I(\{\bold{\psi}(k)\}) = \frac{\min_{\psi(\bold{k})\in\mathcal{C}} Z(k,l)}{\max_{\psi(\bold{k})\in\mathcal{C}} Z(k,l)}, \label{Istr}
\end{equation}
where  $\mathcal{C}=\bold{\Psi}^\top \bold{\Psi}$ is the input correlation matrix of input pattern and $l={1,\ldots, L}$. From \eqref{Istr}, we can see that when the partition function is constant (i.e., stable) for different samples, $I(\{\bold{\psi}(k)\})$ becomes close to $1$ which indicates that the contextual embedding space $\{\bold{\psi}(k)\}$ is more isotropic \cite{arora-etal-2016-latent,mu2018allbutthetop}. 
Note that in \eqref{norm_loglin}, the probability of a value in any time instance is the exponential of the corresponding logit $z_i(k,l)$ divided by the partition function $Z(k,l)$. 
If the partition function remains stable for different samples, the logits can be solely
determined by the probabilities, thereby resolving the shift-invariance problem of the softmax function.

Building on this theoretical foundation, we now turn to the following empirical question: \textit{``How can we measure and interpret isotropy in practice, and how does it relate to generalization across numerical domains?"}. Motivated by Theorem 3.1 and the need for structural constraints in LLM representations, we analyze the effective dimensionality and cluster organization of LLM's hidden representations in the contextual embedding space. These analyses reveal how isotropy manifests in pre-trained LLMs and how its presence correlates with the model’s ability to generalize to time series forecasting across different numerical domains, and hence, provides a performance guarantee. Section~\ref{study} introduces methods for quantifying this internal structural integrity using spectral alignment mechanism, principal component analysis (PCA) and cluster-based isotropy metric, and consequently, linking theoretical reliability (i.e., performance guarantee) to empirical generalizability.

\section {Study of isotropy 
in LLM hidden representations}\label{study}
\vspace{-0.06in}
\paragraph{Analysis Settings.} For this study, we consider five different language models including Chronos-T5~\cite{ansari2024chronos}, Chronos-Bolt\footnote{\url{https://huggingface.co/autogluon/chronos-bolt-base}}, PatchTST~\cite{nie2023a}, Moirai-1.0-R~\cite{woo2024}, and Lag-Llama~\cite{rasul2024}. For illustration, we randomly select a real dataset (i.e., finance-Dataset 1) from a broader collection of 22 numerical datasets that we use in this paper since we see similar results with all of these datasets. The details of these models and datasets could be found in Section~\ref{practical data}.
\vspace{-0.18in}
\begin{wraptable}[7]{r}{9.3cm}
%\begin{table*}[htbp]\label{tab1}
\centering
\caption{The effective dimension $d(0.8)$}
\scalebox{0.9}{
\begin{tabular}{l|cccccccccccc}\label{tab1}
%\hline
Layer & 1 & 2 & 3 & 4 & 5 & 6 & 7 & 8 & 9 & 10 & 11 & 12 \\
\hline
Chronos-T5 & 4 & 4 & 4 & 4 & 4 & 4 & 4 & 4 & 4 & 4 & 4 & 4\\
Chronos-Bolt & 1 & 1 & 1 & 1 & 1 & 1 & 1 & 1 & 1 & 1 & 1 & 1\\
PatchTST & 1 & 1 &  &  &  &  &  &  &  &  &  & \\
Moirai & 1 & 1 & 1 & 1 & 1 & 1 &  &  &  &  &  & \\
Lag-Llma & 2 & 2 & 2 & 2 & 2 & 2 & 2 & 2 &  &  &  & 
\\
\hline
\end{tabular}}
%\end{table*}
\end{wraptable}
\subsection{Effective Dimensions} In each layer of each model, we start with a data matrix $\mathbf{A} \in \mathbb{R}^{|\mathcal{V}| \times D}$, where $|\mathcal{V}|$ represents the number of tokens in the input time series sequence, and $D$ corresponds to the embedding dimension. We apply PCA to reduce
the dimensionality from $D$ to $m$ %PCA reduces this matrix to a more compact form, 
i.e., $\mathbf{\tilde{A}} \in \mathbb{R}^{|\mathcal{V}| \times m}$. Then, 
the fraction of variance captured by the reduced 
representation is given by: $r_m = \frac{\sum_{i=0}^{m-1} \sigma_i}{\sum_{i=0}^{D-1} \sigma_i}$
where $\sigma_i$ denotes the $i$-th largest eigenvalue of the covariance matrix of $\mathbf{A}$. We define the $\epsilon$-effective dimension as $d(\epsilon) \triangleq \arg\min_m r_m \geq \epsilon$.
For instance, if $d(0.8) = 3$, then three principal dimensions retain $80$\% of the variance. A higher $d$ suggests a more isotropic space~\cite{cai2021isotropy}, where information is spread across multiple dimensions rather than being concentrated in a narrow subspace. Table~\ref{tab1} presents the values of $d(0.8)$ for different layers and models. Surprisingly, all of these models have very small effective dimensions as compared to original embedding dimensions. For instance, Chronos-Bolt has very small effective dimensions, with $d(0.8) = 1$ for layers 1 through 12, as compared to its original embedding dimensions $D=512$. The small effective dimensionality is another way of telling that %suggests 
that Chronos-Bolts’s embedding vectors lie in a subspace defined by a very narrow cone \cite{Ethayarajh2019HowCA}, and consequently, their inter-token cosine similarity is large. If all the embedding vectors lie on a $1$-dimensional line, the inter-token cosine similarity would be close to $1$, and there would be hardly any model capacity. Surprisingly, despite having such low effective dimensionality, %majority of their embedding vectors lie in a low-dimensional subspace and are highly similar to one another, 
these language models still perform well in numerical domains.  This counterintuitive result motivate us to look deeper into the contextual embedding space. 
%\vspace{-0.18in}
\subsection{Spectral Alignment for Generalization in Numerical Settings} \label{transport_Example}
%\subsection{Clusters in the Contextual Embedding Space} \label{transport_Example} 
Let $G(\bold{\Psi}) \!= \!(g_1(\bold{\Psi}), \ldots, g_{|\mathcal{V}|}(\bold{\Psi}))^\top \!: \mathbb{R}^{|\mathcal{V}| \times D } \mapsto \mathbb{R}^{|\mathcal{V}| \times D }$ 
be the function for self-attention, i.e., 
$g_i(\bold{\Psi}) = \text{softmax}(\bold{\Psi}\bold{\Lambda}\bold{\Psi}^\top) \bold{\Psi}$, where 
$\bold{\Lambda} \!=\! \bold{W}_Q \bold{W}_K^\top \in \mathbb{R}^{D \times D}$, and $\bold{W}_Q\!\!\in\!\! \mathbb{R}^{D \times m}$, $\bold{W}_K\!\!\in\!\! \mathbb{R}^{D \times m}$ are the parameter matrices for the query and key matrices of self-attention. The lemma below provides insights into how the isotropic property of pre-trained LLMs enables generalization in numerical domains. The proof of this lemma follows the analysis in~\cite{pmlr-v139-kim21i} is provided in Appendix~\ref{lem2} for completeness. 
\vspace{-0.07in}
\begin{lemma}\label{lem1}
Consider the Jacobian matrix $\bold{J} = \left[\frac{\partial g_i(\bold{\Psi})}{\partial \bold{\psi}_j}\right]_{i,j=1}^{|\mathcal{V}|}$, which represents the gradient of the self-attention mapping $G(\bold{\Psi})$ with respect to the input time series token embeddings. Then the spectral norm of $\bold{J}$ satisfies  
$\|\bold{J}\|_2 \leq |\bold{\Lambda}|_2 \sum_{i=1}^{|\mathcal{V}|} \left(p_{i,i} + \frac{1}{2}\right) \left|\bold{\psi}_i - \sum_{j=1}^{|\mathcal{V}|} p_{i,j} \bold{\psi}_j\right|^2 + \Delta$,  
where the residual term $\Delta$ is given by 
$\Delta = |\bold{\Lambda}|_2 \sum_{i \neq j}^{|\mathcal{V}|} p_{i,j} \left|\bold{\psi}_j - \sum_{q=1}^{|\mathcal{V}|} p_{i,q} \bold{\psi}_q\right|^2 + \frac{|\bold{\Lambda}|_2}{2} \sum_{j=1}^{|\mathcal{V}|} |\bold{\psi}_j|^2$,
and the attention weights $p_{i,j}$ are defined as  
$p_{i,j} = \frac{\exp(\bold{\psi}_i^\top \bold{\Lambda} \bold{\psi}_j)}{\sum_{k=1}^{|\mathcal{V}|} \exp(\bold{\psi}_i^\top \bold{\Lambda} \bold{\psi}_k)}$.
\end{lemma}
\vspace{-0.09in}
%Using the gradient structure revealed in Lemma~\ref{lem1}, we can connect self-attention with PCA. 
From Lemma~\ref{lem1}, we can see that, in onder to minimize the norm of the gradient $\|\bold{J}\|_2$, we essentially need to make  $\sum_{i=1}^{|\mathcal{V}|} \left|\bold{\psi}_i - \sum_{j=1}^{|\mathcal{V}|} p_{i,j} \bold{\psi}_j\right|^2$  
small. When $\bold{\Lambda}$ is small and all the input time series token embeddings are centered at the origin, $\sum_{i=1}^{|\mathcal{V}|} \bold{\psi}_i \!=\! 0$, we have  
$\sum_{i=1}^{|\mathcal{V}|} \left|\bold{\psi}_i - \bold{\Psi}^\top p_{i,:}\right|^2 \approx \sum_{i=1}^{|\mathcal{V}|} \left|\bold{\psi}_i - \bold{\Psi}^\top\bold{\Psi} \bold{\Lambda} \bold{\psi}_i\right|^2$ (see Appendix~\ref{lem2}). 

Next, we prove that $\bold{\Lambda}$ minimizes the objective  
$\sum_{i=1}^{|\mathcal{V}|} \left|\bold{\psi}_i - \bold{\Psi}^\top\bold{\Psi} \bold{\Lambda} \bold{\psi}_i\right|^2$ and  
contains the $m$ largest eigenvectors of correlation matrix $\bold{\Psi}^\top \bold{\Psi}$ of time series token embeddings, where $m$ is the rank of $\bold{\Lambda}$. %The proof of Theorem~\ref{th2} is provided in Appendix~\ref{apth2}.
%\vspace{0.1in}
\begin{theorem}\label{th2}
Let the eigenvalues of the correlation matrix $\bold{\Psi}^\top \bold{\Psi}$ be ordered as $\lambda_1 \geq \lambda_2 \geq \cdots \geq \lambda_D$, and let $\gamma_i \in \mathbb{R}^D$ for $i = 1, \ldots, D$ denote their associated eigenvectors. Then, the matrix $\bold{\Lambda}^*$ that minimizes the quantity $\sum_{i=1}^{|\mathcal{V}|} \left|\bold{\psi}_i - \bold{\Psi}^\top\bold{\Psi} \bold{\Lambda} \bold{\psi}_i\right|^2$  
has the optimal form  
$\bold{\Lambda} = \sum_{i=1}^m \frac{1}{\lambda_i} \gamma_i \gamma_i^\top.$
\end{theorem}
\vspace{-0.06in}
\begin{proof} 
The proof of Theorem~\ref{th2} is provided in Appendix~\ref{apth2}.
\end{proof}
Theorem~\ref{th2} shows that %the self-attention learns to perform a function for the numerical downstream tasks through training, which are closely related to the $m$ largest eigenvectors of LLM hidden representations. In other words, 
the self-attention mechanism effectively projects input time series tokens onto a low-dimensional contextual embedding space defined by the top eigenvectors of the correlation matrix $\bold{\Psi}^\top \bold{\Psi}$. This result reveals that the self-attention mechanism in LLMs implicitly aligns with the dominant directions (i.e., top eigenvectors) of the contextual embedding space, and hence, suggesting that isotropy is not just a geometric artifact but a learned structural property that supports effective generalization to numerical downstream tasks.
%\subsubsection{Isotropy within Clusters} 

%\begin{comment}

While the self-attention aligns input representations with the dominant eigenvectors of the embedding space, the alignment may vary across different subregions of the contextual embedding space due to variations in the input sequences, token types, or contextual patterns. As a result, the degree of isotropy may differ across subregions of the contextual embedding space, which motivates the need to assess isotropy at a local (i.e., cluster) level rather than relying solely on a global metric. The next section explores these local structural patterns and examines the geometry of the hidden representations through principal component analysis (PCA), which helps reveal how variance is distributed across embedding dimensions. %and whether distinct clusters emerge. 

\begin{figure}[ht]
%\vspace{-0.08in}
\centering
\includegraphics[width=0.8\linewidth]{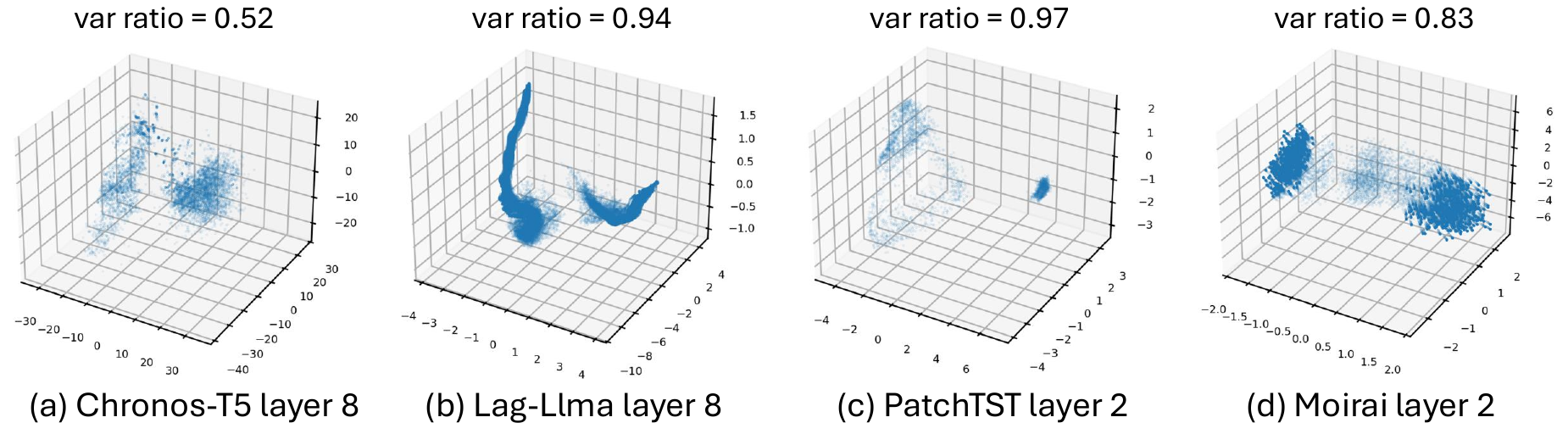}
\caption{Isolated or slightly overlapping cluster islands exist in the contextual embedding space for all models. For brevity, we only show a few representative middle layers from each model.}
\label{pc_plot_GPT2}
%\vspace{-0.14in}
\end{figure}
%\paragraph{Local Structural Analysis of Contextual Embeddings.} 
\subsection{Clusters in the Contextual Embedding Space}
Motivated by the results of Lemma~\ref{lem1} and Theorem~\ref{th2}, this section investigates local structural patterns by projecting the LLMs' hidden representations into a lower-dimensional space using the top $m\!=\!3$ eigenvectors via PCA, as shown in Figure~\ref{pc_plot_GPT2}.
%Motivated by the results from Lemma~\ref{lem1} and  Theorem~\ref{th2}, this section explores the local structural patterns by projecting the models' hidden representations into a lower-dimensional space by using the $m\!=\!3$ largest eigenvectors through PCA, as shown in Figure~\ref{pc_plot_GPT2}. 
The three axes of the figure represent the first three principal components of the covariance matrix of LLM representations of each layer. For instance, in Figure~\ref{pc_plot_GPT2}b and \ref{pc_plot_GPT2}d, the first three principal components account for $94$\% of the total variance in layer 8 of Lag-Llama and $83$\% in layer 2 of Moirai.
%For instance, in layer $8$ of Lag-Llma in Figure~\ref{pc_plot_GPT2} b and in layer $2$ of Moirai in Figure~\ref{pc_plot_GPT2} b, the first three principal components account for $94$\% and $83$\%, respectively, of the total variance. 
From Figure~\ref{pc_plot_GPT2} a, \ref{pc_plot_GPT2} b, \ref{pc_plot_GPT2} c and \ref{pc_plot_GPT2} d, we can see that there are disconnected or slightly overlapping islands that are far away from each other. %Note that the first principal dimension value spans from $0$ to $2000$, significantly wider than the other $2$ dimensions, and dominates the total variance. A similar analogy can also be observed for transport Dataset 2 in Figure~\ref{pc_plot_GPT2} b. 
In \eqref{cosine_nocluster}, the space isotropy is measured on pairs of arbitrary time series token representations, which could reside in two disconnected clusters. However, given that the variance is dominated by distances between clusters, such estimation would be biased by the inter-cluster distances. Hence, it is more reasonable to consider a per-cluster (i.e., local) investigation rather than a global estimate.

\paragraph {Isotropy within Clusters.} We start by performing clustering on the LLM representations in the contextual embedding space. There are various methods for performing clustering, such as $K$-means and DBSCAN algorithm~\cite{DBS}. We select $K$-means clustering method because it is reasonably fast in high embedding dimensions. %(e.g., $d\geq768$ for GPT2, ELMo, BERT etc.). 
We use the classical silhouette score analysis \cite{ROUSSEEUW198753} to determine the number of clusters $|C|$ in the contextual embedding space (see Appendix~\ref{cluster_appendix} for details).
%\textbf{Isotropy within Clusters.} 
Since each LLM contextual embedding instance $\bold{\psi}(k_i)$ belongs to a particular cluster through clustering, the cosine similarity should be measured after shifting the mean to the origin \cite{mu2018allbutthetop}. Accordingly, we subtract the mean for each cluster (i.e., centroid) and calculate the adjusted $\zeta_{\text{cos}}$ in Section~\ref{model}. Assuming we have a total of $|C|$ clusters, let $\bold{\psi}_c(k_i) = \{\bold{\psi}_c^1(k_i), \bold{\psi}_c^2(k_i), \dots\}$ be the set of token $k$’s contextual embeddings in cluster $c \in C$, and $\bold{\psi}_c(k_i)$ be one random sample in $\bold{\psi}_c(k_i)$. We define the adjusted inter-token cosine similarity as
%\vspace{-0.05in}
\begin{align}\label{ad_cos_sim}
\zeta'_{\textrm{cos}} \triangleq \mathbb{E}_c \left[ \mathbb{E}_{i \neq j} \left[ \cos \left( \bar{\bold{\psi}}_c(k_i), \bar{\bold{\psi}}_c(k_j) \right) \right] \right],
\vspace{-0.18in}
\end{align}
where $\bar{\bold{\psi}}_c(k_i) = \bold{\psi}_c(k_i) - \mathbb{E}_{\bold{\psi}_c} [\bold{\psi}_c(k_i)]$. Here $\mathbb{E}_c$ is the average over different clusters, and $\bar{\bold{\psi}}_c(k_i)$ is the original contextual embedding shifted by the mean, with the mean taken over the samples in cluster $c$ \cite{pmlr-v139-kim21i}. The inter-token cosine similarity takes values between $-1$ and $1$. A value close to 0 indicates strong isotropy and ensures the existence of structure in the LLM representations.

To put it in a nutshell, this section provides a theoretical foundation showing that self-attention projects input tokens onto a low-dimensional subspace aligned with the dominant eigenvectors of the embedding correlation matrix. This alignment induces isotropy in LLM hidden representations, stabilizing the partition function and preserving the structure needed for reliable numerical downstream task performances. 
In Section~\ref{practical data}, we extend this analysis by empirically evaluating how isotropy in different language models' hidden representations correlates with time series forecasting performances across a wide range of numerical datasets, varying context lengths, and noise levels.
%\vspace{-0.16in}
\section{Experiments}\label{practical data}
\vspace{-0.09in}
\textbf{Baselines.} We consider popular pre-trained LLMs as the baselines for numerical downstream tasks, including Chronos-T5~\cite{ansari2024chronos} and Choronos-Bolt~(https://huggingface.co/autogluon/chronos-bolt-base), PatchTST~\cite{nie2023a}, Moirai-1.0-R~\cite{woo2024} and Lag-Llama~\cite{rasul2024}. 
%\vspace{-0.1in}
%%%%%%%%%%%%%%%%%%%%%%%%%%%%%%%%%%
\begin{table}[ht!]
\centering
\caption{LLM models architectures, time series tokenization techniques and hyperparameter choices. %Hyperparameters not specified are set to defaults in their respective implementations. 
$L$ stands for context length, $d_h$ for hidden layer dimension, $n_L$ for number of layers, $n_H$ for number of heads, and $\eta$ for learning rate.}
\vspace{0.1in}
\label{tab:all-baselines}
\scalebox{0.8}{
\begin{tabular}{l l l p{7.8cm}}
\toprule
\textbf{Model} & \textbf{Architecture} & \textbf{Tokenization Technique} & \textbf{Hyperparameters} \\
\midrule
Chronos-T5       &  \parbox{3.9cm}{Encoder-Decoder with\\ autoregressive forecasting}    & Scaling \& Quantization & Default \\
\midrule
Chronos-Bolt  & \parbox{3.7cm}{Encoder-Decoder with\\ multi-step forecasting}   & Scaling \& Quantization  & Default\\
\midrule
PatchTST       & Vanilla Encoder  & Patching & Patch length: 16, Stride: 8, $d_h = 32, n_L = 2, n_{H} = 4$   \\
\midrule
Moirai   & Encoder & Patching &  $L = 1024$, Patch length: selected by dataset-specific validation     \\
\midrule
%LLM4CP &  Task-specific& & Default
Lag-Llama    & Decoder  & Lag Feature &     $L = 32$  
\\
\bottomrule
\end{tabular}}
%%%%%%%%%%%%%%%%%%%%%%%%%%%
%\vspace{-0.3in}
\end{table}
The considered models use different architectures, time series tokenization techniques and hyperparameters for numerical downstream tasks. 
For instance, Lag-Llma  use decoder only transformer, PatchTST and Moirai-1.0-R use vanilla Transformer enoder, while Chronos-T5 and Choronos-Bolt use encoder-decoder transformer. Different baselines achieve contextual embedding in different ways. For example, PatchTST focuses on tokenizing time series as patches and uses self-attention for modeling dependencies within each patch and across patches, while CHORNOS-T5 and CHRONOS-Bolt adapt language modeling architectures minimally and generate categorical tokens by applying scaling and quantization. The details of these baselines are summarized in Table~\ref{tab:all-baselines}.

\begin{table}[htb!]
    \centering
    \caption{Real and Synthetic Datasets}
    \scalebox{0.8}{
    \begin{tabular}{l|l|l|l}
        \toprule
        \textbf{Data Subset} & \textbf{Domain} &\textbf{Dataset 1} & \textbf{Dataset 2} \\
        \midrule
        \multirow{6}{*}{Real Datasets} 
        & Energy & Australian Electricity – Queensland State 
        & Australian Electricity – South Australia \\
        & Weather & Solar Radiation & Rainfall \\
        & Finance & Exchange Rate & NN5 Weekly Cash Withdrawals \\
        & Healthcare & Hospital Patient Counts & COVID-19 Deaths \\
        & Transportation & Transportation Signaling 1 & Transportation Signaling 2 \\
        & Retail & Car Sales & Dominick \\
        \midrule
        \multirow{5}{*}{Synthetic Datasets} 
        & Linear & DotProduct kernel (C=0) & DotProduct kernel (C=1) \\
        & seasonality &seasonality kernel (period = 0.5W) & seasonality kernel (period = 0.25H) \\
        & Trend &RationalQuadratic kernel  ($\alpha=1$) & RationalQuadratic kernel ($\alpha=10$) \\ 
        & Non-Linear & RBF kernel (length scale $= 0.1$) & RBF kernel (length scale $= 1 $)\\
        & Stochastic & WhiteKernel (noise level = 0.1) & WhiteKernel (noise level = 1) \\
        \bottomrule
    \end{tabular}}
    \label{tab:data_subsets_datasets}
    \vspace{-0.1in}
\end{table}
\textbf{Datasets.} We conduct a comprehensive evaluation using $12$ different real time series datasets from various numerical domains, including energy, nature, finance, healthcare, retail and transportation. The sources of these open-source datasets along with their descriptions, including how each dataset is used across different LLM can be found in Table~\ref{tab:all-datasets} of Appendix~\ref{data_des}. We also illustrate our findings using KernelSynth~\cite{ansari2024chronos} (see  Algorithm~\ref{alg1} in Appendix~\ref{data_des} for details), a method that generates $10$ additional synthetic datasets via Gaussian processes in Section~\ref{practical data}. 
%The detailed description of both real and synthetic datasets is provided in Appendix~\ref{data_des}. 
We select two different datasets from each numerical domain (as shown in Table~\ref{tab:data_subsets_datasets}) and then perform qualitative analysis with synthetic datasets and quantitative analysis with real datasets. The results of these analyses are provided in the next two sections. %in Appendix~\ref{data_des}.
%\vspace{-0.09in}
%\vspace{-0.09in}
\subsection{Qualitative Analysis}\label{quality}
\vspace{-0.1in}
We now analyze the time series forecasting by the baseline LLMs qualitatively. We focus on synthetically
generated time series for a controlled analysis 
of different types of time series patterns which belong to $5$ different domains, such as linear, seasonality, trend, non-linear and stochastic. 
We are particularly interested in the isotropic measurement in the LLM's last layer as it is related to the logits and probabilistic inference as explained in Section~\ref{model}. So all isotropic measure provided in this section is based on the last layer of the baselines.
%\vspace{-0.05in}
\begin{figure}[ht]
\centering
\includegraphics[width=1\linewidth]{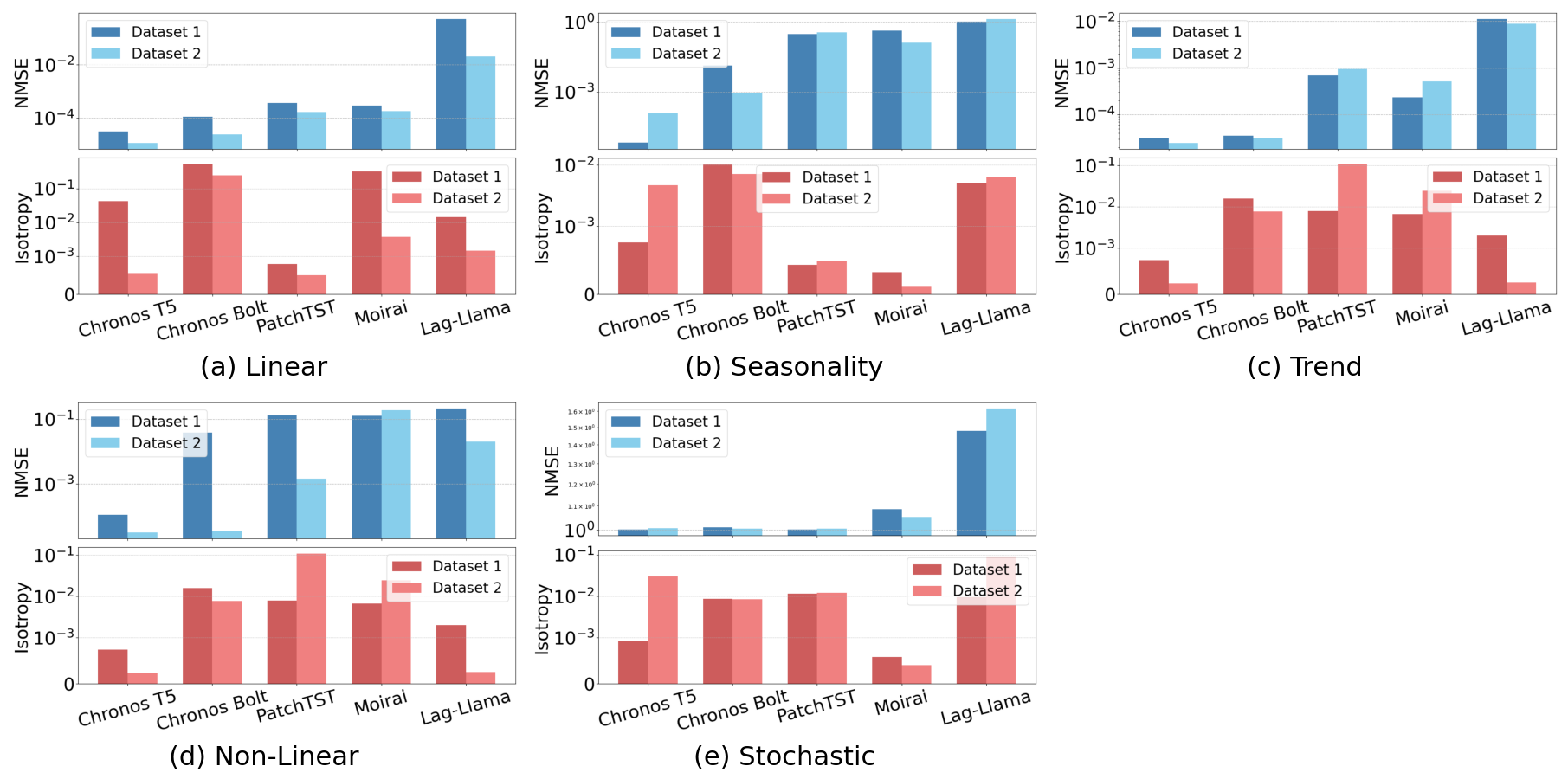}
\vspace{-0.15in}
\caption{NMSE vs isotropy analysis for $10$ different synthetic datasets of $5$ different domains. %The inter-token cosine similarity value close to zero means strong isotropy and vice versa.
}
\label{Syn_cos}
\vspace{-0.18in}
\end{figure}

We begin by analyzing time series forecasting performance (i.e., NMSE) for different baselines and its relation with isotropy in Figure~\ref{Syn_cos}. 
For instance, in Figure~\ref{Syn_cos} b, we have (NMSE = $0.0000066$ and cosine similarity = $|-0.00076|$) for seasonality-Dataset 1 and (NMSE = $0.00012$ and cosine similarity = $0.0047$) for seasonality-Dataset 2 for Chronos-T5. This shows that stronger isotropy exists (i.e., inter-token cosine similarity value is close to $0$) in Chronos-T5's embedding space 
for seasonality-Dataset 1 which preserves the structure in its hidden representations and causes good downstream task performance. 
On the other hand, a weaker isotropy exists (i.e., inter-token cosine similarity value is far from $0$) in Chronos-T5's embedding space 
for seasonality-Dataset 2, which, in turn, causes a lack of structure in its hidden representations, thereby leading to bad forecasting performance as compared to seasonality-Dataset 1. %The NMSE and inter type cosine similarity may also vary for different language models and datasets. For example, in Figure~\ref{Syn_cos} c, we can see that thethe NMSE for trend-Dataset 1 is lower for PatchTST and Moirai,  while higher for Chronos-T5, Chronos-Bolt and Lag-Llma as compared to trend-Dataset 2. In contrast, the NMSE for trend-Dataset 2 is lower for Chronos-T5, Chronos-Bolt, and Lag-Llma, while higher for PatchTST and Moirai as compared to trend-Dataset 1. 
The NMSE and inter-type cosine similarity can also vary across different language models and datasets. 
For example, in Figure~\ref{Syn_cos}c, the NMSE for trend-Dataset 1 is lower for PatchTST and Moirai, but higher for Chronos-T5, Chronos-Bolt, and Lag-Llama, compared to their respective NMSE on trend-Dataset 2. Conversely, for trend-Dataset 2, the NMSE is lower for Chronos-T5, Chronos-Bolt, and Lag-Llama, but higher for PatchTST and Moirai, compared to their respective NMSE on trend-Dataset 1. A similar analysis can also be observed for other synthetic datasets and baselines in Figures~\ref{Syn_cos} b, \ref{Syn_cos} d, and \ref{Syn_cos} e.
\begin{figure}[ht]
\centering
\includegraphics[width=0.6\textwidth]{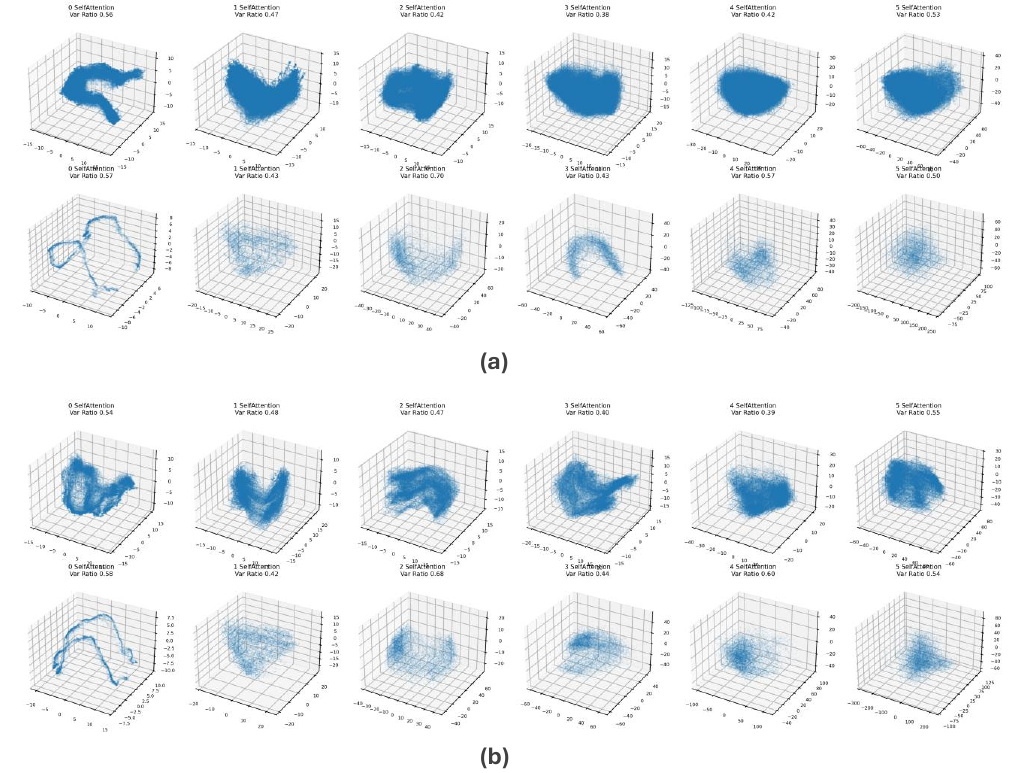}
%\vspace{-10pt}
\caption{Variations in Chronos-T5's hidden representations for different input context lengths for the same synthetic dataset non-linear-Dataset 1
 : (a) Contextual embedding space for input context length $ L = 500$. (b) Contextual embedding space for input context length $L = 100$. 
}
\label{pca_plot_context}
\vspace{-0.1in}
\end{figure}
This shows that any dataset from any particular domain may cause different forecasting performances for different baselines, as it generates different hidden representations (see Appendix~\ref{full_pca_plots} for full visualization) in contextual embedding spaces, and hence, different isotropy measures. %, for language models with different architectures and tokenization techniques.  
%\vspace{-0.1in}

Next, we examine the influence of isotropy on forecasting performance in two important scenarios: a) different input context lengths, and b) different levels of noises in the input data.
The first scenario is important as it provides an analysis that helps guide in selecting reasonable input context lengths rather than selecting the length through random trials and errors. 
The second scenario is important as it gives us ideas on how the level of noise in noisy data impacts performance, since the data in the real world is mostly noisy.

\textbf{Isotropy in different input context lengths.} We first analyze the effect of isotropy under varying input context lengths. 
We begin with an illustration in Figure~\ref{pca_plot_context} %in Appendix~\ref{full_pca_plots}, 
where we show how the hidden representations of Chronos-T5 vary for two different input context lengths,
such as $L=500$ and $L=100$, for non-linear-Dataset 1, 
%\clearpage
which generates different isotropic measures for different input context lengths. %We use Choronos-T5 as an example model and non-linear-Dataset 1 as an example dataset for this illustration.

%%%%%%%%%%%%%%%%%%%%
\begin{wrapfigure}[22]{r}{0.48\textwidth}
\centering
\includegraphics[width=.4\textwidth]
{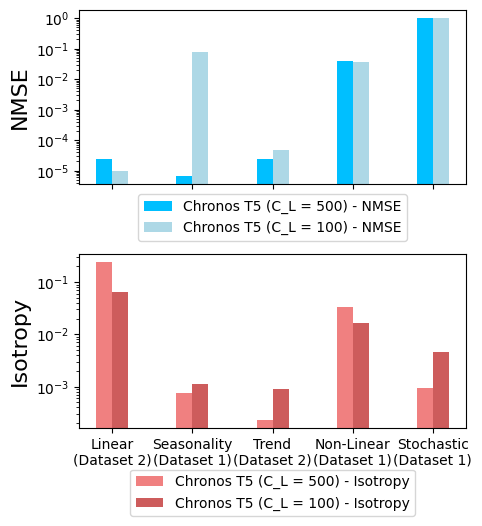}
%\vspace{-0.18in}
\caption{NMSE vs isotropy comparison across different input context lengths for synthetic datasets.}
\label{T5_context}
\vspace{-0.14in}
\end{wrapfigure}
In Figure~\ref{T5_context}, we compare the NMSE vs isotropy across two different input context lengths, $L=500$ and $L=100$, for different synthetic datasets. We use Choronos-T5 as an example model for this experiment. As can be seen from the figure, the isotropy values vary across different input context lengths and datasets. 
For instance, in seasonality-Dataset 1, we have (NMSE$=0.0000066$, cosine similarity$=|-0.00076|$) and (NMSE$=0.0793$, cosine similarity$=0.0011$) for $L=500$ and $L=100$, respectively. 
The decrease in isotropy significantly increases the NMSE for the input context length $L=100$. In contrast, in linear-Dataset 2,  we have (NMSE$=0.000025$, cosine similarity$=0.2474$) and (NMSE$=0.000009$, cosine similarity$=0.0644$) for $L=500$ and $L=100$, respectively. In this scenario, the isotropy increases for the input context length $L=100$, which causes the decrease in NMSE for chornos-T5.
In practice, the input context length is often selected randomly or through trial and error, which may cause higher forecasting errors for different datasets. 
Isotropy analysis enables us to understand 
how varying input context lengths influence the hidden representations of the language 
model. 
This insight helps guide improvements in forecasting performance by examining the isotropic properties of the contextual embedding space.

\begin{wrapfigure}[17]{r}
{0.48\textwidth}
%\vspace{-0.3in}
\centering
\includegraphics[width=.4\textwidth]
{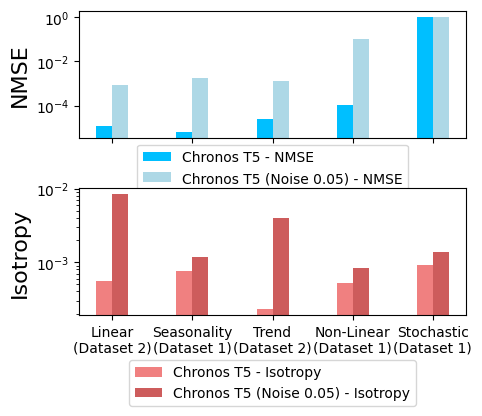}
\vspace{-0.1in}
\caption{NMSE vs isotropy comparison across different noise levels in synthetic datasets.}
\label{T5_noise}
%\vspace{-0.3in}
\end{wrapfigure}
\textbf{Isotropy in varying noise levels in datasets.} Next, we focus on
the second scenario to see the impact
of noisy datasets on LLM's performance. 
Again, we use the Chornos-T5 as an example language model. 
Figure~\ref{T5_noise} compares the NMSE vs isotropy across two different cases, one without noise, and the other with Gaussian noise with a standard deviation $\sigma=0.05$ standard deviation.
From Figure~\ref{T5_noise}, we can see consistently lower isotropy (i.e., inter-token cosine similarity far from $0$) for all noisy synthetic datasets as compared to the datasets without noise.
For instance, in trend-Dataset 2, we have (NMSE$=0.000024$, cosine similarity$=|-0.00022|$) and (NMSE$=0.0012$, cosine similarity$=0.0040$) for $\sigma=0$ and $\sigma=0.05$, respectively. The decrease in isotropy significantly increases the NMSE for the noisy dataset. In practice, many real-world numerical domains—such as those in nature and energy—exhibit noisy and dynamic behavior. 
\begin{figure}[ht]
\vspace{-0.095in}
\centering
\includegraphics[width=1\linewidth]{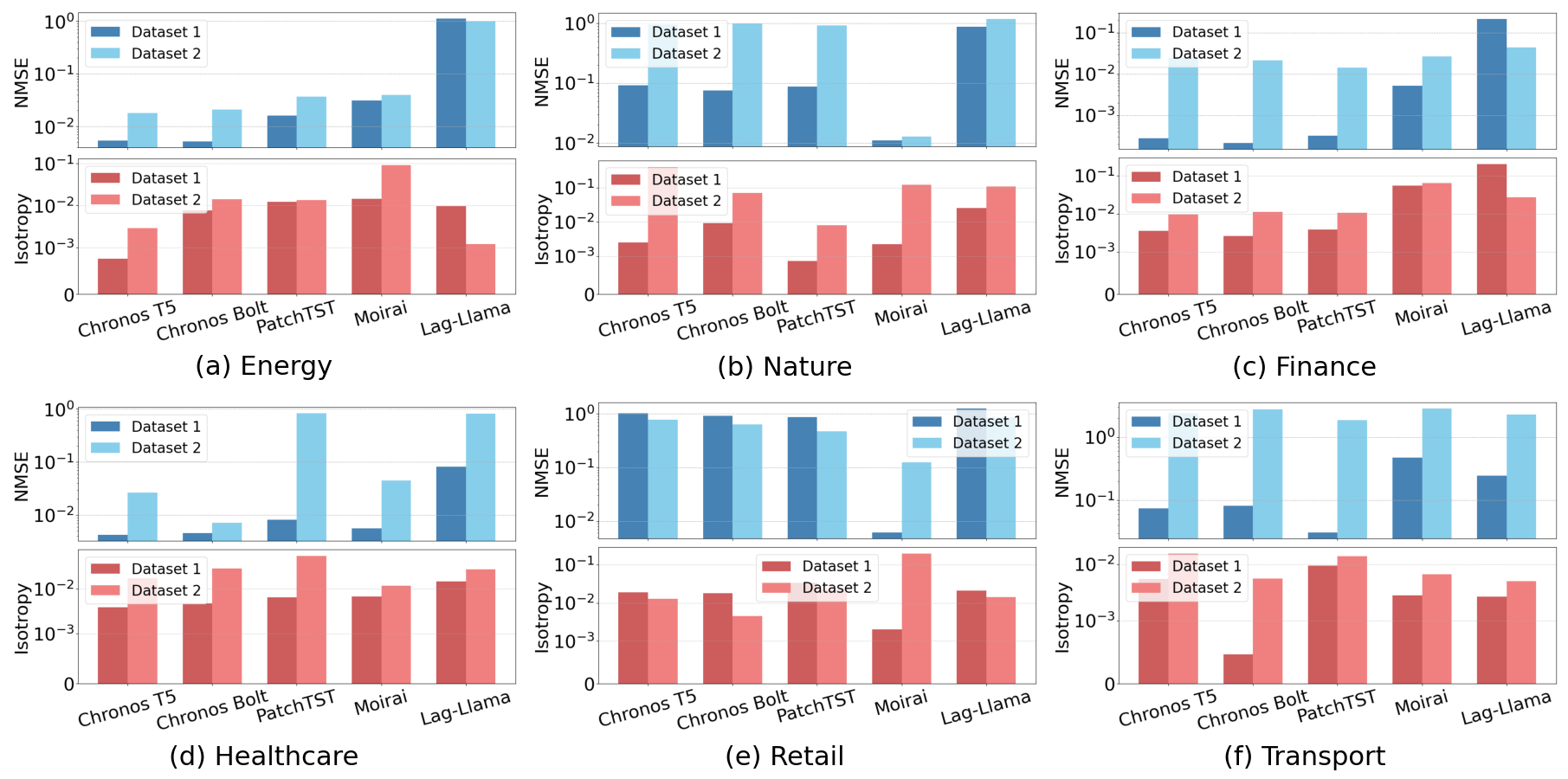}
\vspace{-0.2in}
\caption{NMSE vs isotropy analysis for $12$ different real datasets of $6$ different domains. %The inter-token cosine similarity value close to zero means strong isotropy and vice versa.
}
\label{real_cos}
\vspace{-0.16in}
\end{figure}
In these environments, it is often infeasible to measure noise in real time or to pre-process the input time series for improved performance. However, the isotropy in the hidden representations of LLMs can be readily measured, and thus, can be leveraged to enhance forecasting performance by identifying and mitigating the effects of noisy inputs in contextual embedding space.
%In practice, the environments of many real numerical domains, such as nature and energy, are noisy and dynamic. In such an environment, it is not always possible to measure the noise in real time and take the necessary steps to clean the input time series for better performance. However, it is always possible to measure the isotropy from LLM hidden representations, which can be used as a measure of noise in the input datasets, and thus, help to improve the forecasting performance.    
%\vspace{-0.12in}
\subsection{Quantitative Analysis}
\vspace{-0.33\in}
Next, we present our main results on $12$ real datasets which belong to $6$ different numerical domains including energy, nature, finance, healthcare, retail, and transportation.
As our qualitative analysis in Section~\ref{quality}, we select two different datasets from each numerical domain and the isotropy measure from LLM's last layer to show the impact of isotropy on NMSE performance for different language models. 
%\vspace{-0.09in}

In Figure~\ref{real_cos}, we analyze the time series forecasting performance of different baselines and its relation with isotropy for different real datasets. For instance, in Figure~\ref{real_cos} e, 
we have (NMSE = $0.0061$ and cosine similarity = $0.0020$) for retail-Dataset 1 and (NMSE = $0.1255$ and cosine similarity = $0.1931$) for retail-Dataset 2 for Moirai. 
This indicates the existence of stronger isotropy in Moirai's embedding space for retail-Dataset 1 which preserves the structure in its hidden representations and causes good downstream task performance. 
On the other hand, %we have (NMSE = $0.00012$ and cosine similarity = $0.0047$)  for Chronos-T5 and seasonality-Dataset 2. This indicates 
a weaker isotropy exists in Moirai's embedding space for retail-Dataset 2, which yields a lack of structure in its hidden representations and, consequently, bad downstream task performance as compared to retail-Dataset 1. %The NMSE and inter type cosine similarity may also vary for different real datasets and language models. For example, in Figure~\ref{real_cos} a, we can see that the NMSE for energy-Dataset 1 is lower for Chronos-T5, Chronos-Bolt, PatchTST and Moirai,  while higher Lag-Llma as compared to energy-Dataset 2. In contrast, the NMSE for energy-Dataset 2 is lower for Moirai while higher for other baselines as compared to energy-Dataset 1.
The NMSE and inter-type cosine similarity can vary across different real datasets and language models. For example, in Figure~\ref{real_cos}a,
the NMSE for energy-Dataset 1 is lower for Chronos-T5, Chronos-Bolt, 
PatchTST, and Moirai, but higher for Lag-Llama, 
compared to their respective NMSE on energy-Dataset 2. 
Conversely, the NMSE for energy-Dataset 2 is 
\begin{wrapfigure}[19]{r}{0.48\textwidth}
%\begin{figure}
\centering
\includegraphics[width=.4\textwidth]
{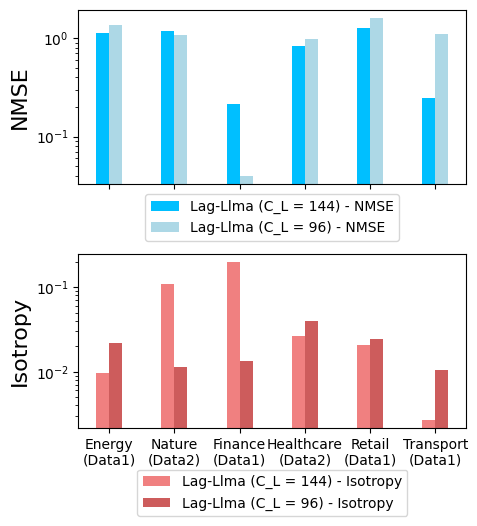}
\vspace{-0.1 in}
\caption{NMSE vs isotropy comparison across different input context lengths for real datasets.}
\label{Lag_context}
\vspace{-0.14in}
\end{wrapfigure}
lower for Moirai but higher for the other baselines, compared to their respective NMSE on energy-Dataset 1. 
A similar analysis can also be observed for other synthetic datasets and baselines in  Figure~\ref{real_cos} c and \ref{real_cos} e. This again shows that datasets from the same numerical domain can cause varying forecasting performance across different baselines, 
as they generate distinct hidden representations in contextual embedding spaces, and hence, different isotropy measures, depending on the language model architecture and tokenization strategy.

Finally, in Figure~\ref{Lag_context}, we compare the NMSE vs isotropy for varying input context lengths to observe its
impact on the real datasets. We select Lag-Llma as our example model. We compare the results for two different input context lengths: 1) the recommended input context 
length $L=144$ and the reduced input context length $L=96$. As can be seen from the figure,  the inter-token cosine similarity values become far from $0$, i.e., from $0.0097$ to $0.0220$ for energy-Dataset 1 and from $0.0026$ to $0.0103$ for transport-Dataset 1, which in turn decreases the NMSE performances.
On the other hand, the inter-token cosine similarity values become close to $0$, i.e., from $0.1091$ to $0.0112$ for nature-Dataset 2 and from $0.2014$ to $0.0133$ for finance-Dataset 1, which in turn improves the NMSE performances. Thus, the variation in the recommended input context length 
may not only decrease the NMSE performances, but can also increase for some datasets.

\section{Conclusion and Limitations}
\vspace{-0.1in}
\label{conclution}
%Isotropy in embeddings as studied here can serve as a foundation for future research on the deeper understanding of LLMs and their applications in various domains. Beyond isotropy, there could be other methods to approximate the partition function with a constant and make the logits useful for the numerical downstream tasks. Moreover, our isotropy study only ensured the existence of structure in the LLM hidden representations and provides a performance guarantee when the structure is preserved by isotropy. Improving the numerical downstream task performance when structure is not preserved in the LLM representations is a topic of future work. 
In this work, we introduced a novel approach to investigate the role of isotropy in LLM hidden representations for numerical downstream tasks.
By deriving an upper bound for the Jacobian matrix which collects all first-order partial derivatives of self-attention with respect to the input pattern, we showed that the self-attention mechanism implicitly aligns with the dominant eigenvectors of the input correlation structure and induces isotropy in the contextual embedding space. The existence of isotropy in the contextual embedding space was found to stabilize the partition function and enable better generalization in numerical downstream tasks across different models and datasets. Our empirical analysis across $10$ synthetic and $12$ real numerical datasets, and $5$ different language models further validated the consistent relationship between isotropy and forecasting performance, highlighting isotropy as a reliable indicator of structured representation learning. These insights open up a new interpretability frontier for LLMs in numerical domains.

While isotropy offers a principled way to preserve useful structure, there may be alternative approaches to approximating the partition function and guiding numerical reasoning. Moreover, developing mechanisms to recover or enhance structure when isotropy is weak remains an important avenue for future work. Ultimately, we believe that incorporating structural insights like isotropy into the LLM design pipeline can significantly improve their reliability and adaptability to numerical domains.

\clearpage

\bibliography{example_paper}

\begin{thebibliography}{24}
\providecommand{\natexlab}[1]{#1}
\providecommand{\url}[1]{\texttt{#1}}
\expandafter\ifx\csname urlstyle\endcsname\relax
  \providecommand{\doi}[1]{doi: #1}\else
  \providecommand{\doi}{doi: \begingroup \urlstyle{rm}\Url}\fi

\bibitem[Ansari et~al.(2024)Ansari, Stella, Turkmen, Zhang, Mercado, Shen, Shchur, Rangapuram, Pineda~Arango, Kapoor, Zschiegner, and et~al.]{ansari2024chronos}
Abdul~Fatir Ansari, Lorenzo Stella, Caner Turkmen, Xiyuan Zhang, Pedro Mercado, Huibin Shen, Oleksandr Shchur, Syama~Syndar Rangapuram, Sebastian Pineda~Arango, Shubham Kapoor, Jasper Zschiegner, and Maddix et~al.
\newblock Chronos: Learning the language of time series.
\newblock \emph{arXiv preprint arXiv:2403.07815}, 2024.

\bibitem[Arora et~al.(2016)Arora, Li, Liang, Ma, and Risteski]{arora-etal-2016-latent}
Sanjeev Arora, Yuanzhi Li, Yingyu Liang, Tengyu Ma, and Andrej Risteski.
\newblock A latent variable model approach to {PMI}-based word embeddings.
\newblock volume~4, pages 385--399, Cambridge, MA, 2016. MIT Press.
\newblock \doi{10.1162/tacl_a_00106}.

\bibitem[Borisov et~al.(2023)Borisov, Sessler, Leemann, Pawelczyk, and Kasneci]{borisov2023language}
Vadim Borisov, Kathrin Sessler, Tobias Leemann, Martin Pawelczyk, and Gjergji Kasneci.
\newblock Language models are realistic tabular data generators.
\newblock In \emph{The Eleventh International Conference on Learning Representations}, 2023.

\bibitem[Cai et~al.(2021)Cai, Huang, Bian, and Church]{cai2021isotropy}
Xingyu Cai, Jiaji Huang, Yuchen Bian, and Kenneth Church.
\newblock Isotropy in the contextual embedding space: Clusters and manifolds.
\newblock In \emph{International Conference on Learning Representations}, 2021.

\bibitem[Dinh et~al.(2022)Dinh, Zeng, Zhang, Lin, Gira, Rajput, Sohn, Papailiopoulos, and Lee]{dinh2022lift}
Tuan Dinh, Yuchen Zeng, Ruisu Zhang, Ziqian Lin, Michael Gira, Shashank Rajput, Jy-yong Sohn, Dimitris Papailiopoulos, and Kangwook Lee.
\newblock Lift: Language-interfaced fine-tuning for non-language machine learning tasks.
\newblock \emph{Advances in Neural Information Processing Systems}, 35:\penalty0 11763--11784, 2022.

\bibitem[Dooley et~al.(2023)Dooley, Khurana, Mohapatra, Naidu, and White]{dooley2023}
Samuel Dooley, Gurnoor~Singh Khurana, Chirag Mohapatra, Siddartha Naidu, and Colin White.
\newblock Forecastpfn: Synthetically-trained zero-shot forecasting, 2023.
\newblock URL \url{https://arxiv.org/abs/2311.01933}.

\bibitem[Ester et~al.(1996)Ester, Kriegel, Sander, and Xu]{DBS}
Martin Ester, Hans-Peter Kriegel, J\"{o}rg Sander, and Xiaowei Xu.
\newblock A density-based algorithm for discovering clusters in large spatial databases with noise.
\newblock In \emph{Proceedings of the Second International Conference on Knowledge Discovery and Data Mining}, KDD'96, page 226–231. AAAI Press, 1996.

\bibitem[Ethayarajh(2019)]{Ethayarajh2019HowCA}
Kawin Ethayarajh.
\newblock How contextual are contextualized word representations? comparing the geometry of bert, elmo, and gpt-2 embeddings.
\newblock In \emph{Conference on Empirical Methods in Natural Language Processing}, 2019.

\bibitem[Gao et~al.(2024)Gao, Bai, Gu, Xia, Torr, Li, and Liu]{gao2024inducing}
Kuofeng Gao, Yang Bai, Jindong Gu, Shu-Tao Xia, Philip Torr, Zhifeng Li, and Wei Liu.
\newblock Inducing high energy-latency of large vision-language models with verbose images.
\newblock In \emph{ICLR}, 2024.

\bibitem[Garza and Mergenthaler-Canseco(2023)]{garza2023timegpt}
Azul Garza and Max Mergenthaler-Canseco.
\newblock Timegpt-1.
\newblock \emph{arXiv preprint arXiv:2310.03589}, 2023.

\bibitem[Gruver et~al.(2024)Gruver, Finzi, Qiu, and Wilson]{gruver2024}
Nate Gruver, Marc Finzi, Shikai Qiu, and Andrew~Gordon Wilson.
\newblock Large language models are zero-shot time series forecasters, 2024.
\newblock URL \url{https://arxiv.org/abs/2310.07820}.

\bibitem[Jin et~al.(2024)Jin, Wang, Ma, Chu, Zhang, Shi, Chen, Liang, Li, Pan, and Wen]{jin2023time}
Ming Jin, Shiyu Wang, Lintao Ma, Zhixuan Chu, James~Y. Zhang, Xiaoming Shi, Pin-Yu Chen, Yuxuan Liang, Yuan-Fang Li, Shirui Pan, and Qingsong Wen.
\newblock Time-{LLM}: Time series forecasting by reprogramming large language models.
\newblock 2024.

\bibitem[Kim et~al.(2021)Kim, Papamakarios, and Mnih]{pmlr-v139-kim21i}
Hyunjik Kim, George Papamakarios, and Andriy Mnih.
\newblock The lipschitz constant of self-attention.
\newblock In Marina Meila and Tong Zhang, editors, \emph{Proceedings of the 38th International Conference on Machine Learning}, volume 139 of \emph{Proceedings of Machine Learning Research}, pages 5562--5571. PMLR, 18--24 Jul 2021.

\bibitem[Liu et~al.(2024)Liu, Liu, Gao, Cheng, and Yang]{liu2024llm4cp}
Boxun Liu, Xuanyu Liu, Shijian Gao, Xiang Cheng, and Liuqing Yang.
\newblock Llm4cp: Adapting large language models for channel prediction.
\newblock \emph{Journal of Communications and Information Networks}, 9\penalty0 (2):\penalty0 113--125, 2024.
\newblock \doi{10.23919/JCIN.2024.10582829}.

\bibitem[Mu and Viswanath(2018)]{mu2018allbutthetop}
Jiaqi Mu and Pramod Viswanath.
\newblock All-but-the-top: Simple and effective postprocessing for word representations.
\newblock In \emph{International Conference on Learning Representations}, 2018.

\bibitem[Nie et~al.(2023)Nie, Nguyen, Sinthong, and Kalagnanam]{nie2023a}
Yuqi Nie, Nam~H Nguyen, Phanwadee Sinthong, and Jayant Kalagnanam.
\newblock A time series is worth 64 words: Long-term forecasting with transformers.
\newblock In \emph{The Eleventh International Conference on Learning Representations}, 2023.

\bibitem[Rasul et~al.(2024)Rasul, Ashok, Williams, Ghonia, Bhagwatkar, Khorasani, Bayazi, Adamopoulos, Riachi, Hassen, Biloš, Garg, Schneider, Chapados, Drouin, Zantedeschi, Nevmyvaka, and Rish]{rasul2024}
Kashif Rasul, Arjun Ashok, Andrew~Robert Williams, Hena Ghonia, Rishika Bhagwatkar, Arian Khorasani, Mohammad Javad~Darvishi Bayazi, George Adamopoulos, Roland Riachi, Nadhir Hassen, Marin Biloš, Sahil Garg, Anderson Schneider, Nicolas Chapados, Alexandre Drouin, Valentina Zantedeschi, Yuriy Nevmyvaka, and Irina Rish.
\newblock Lag-llama: Towards foundation models for probabilistic time series forecasting, 2024.
\newblock URL \url{https://arxiv.org/abs/2310.08278}.

\bibitem[Rousseeuw(1987)]{ROUSSEEUW198753}
Peter~J. Rousseeuw.
\newblock Silhouettes: A graphical aid to the interpretation and validation of cluster analysis.
\newblock \emph{Journal of Computational and Applied Mathematics}, 20:\penalty0 53--65, 1987.
\newblock ISSN 0377-0427.

\bibitem[Wang and Zhang(2024)]{wang2024large}
Dandan Wang and Shiqing Zhang.
\newblock Large language models in medical and healthcare fields: applications, advances, and challenges.
\newblock \emph{Artificial Intelligence Review}, 57\penalty0 (299):\penalty0 1--27, 2024.

\bibitem[Wei et~al.(2021)Wei, Xie, and Ma]{NEURIPS2021_86b3e165}
Colin Wei, Sang~Michael Xie, and Tengyu Ma.
\newblock Why do pretrained language models help in downstream tasks? an analysis of head and prompt tuning.
\newblock In M.~Ranzato, A.~Beygelzimer, Y.~Dauphin, P.S. Liang, and J.~Wortman Vaughan, editors, \emph{Advances in Neural Information Processing Systems}, volume~34, pages 16158--16170. Curran Associates, Inc., 2021.
\newblock URL \url{https://proceedings.neurips.cc/paper_files/paper/2021/file/86b3e165b8154656a71ffe8a327ded7d-Paper.pdf}.

\bibitem[Woo et~al.(2024)Woo, Liu, Kumar, Xiong, Savarese, and Sahoo]{woo2024}
Gerald Woo, Chenghao Liu, Akshat Kumar, Caiming Xiong, Silvio Savarese, and Doyen Sahoo.
\newblock Unified training of universal time series forecasting transformers, 2024.
\newblock URL \url{https://arxiv.org/abs/2402.02592}.

\bibitem[Wu et~al.(2023)Wu, Lee, and Ge]{wu2023connecting}
Chenwei Wu, Holden Lee, and Rong Ge.
\newblock Connecting pre-trained language model and downstream task via properties of representation.
\newblock In \emph{Thirty-seventh Conference on Neural Information Processing Systems}, 2023.

\bibitem[Xu et~al.(2024)Xu, Kurisummoottil~Thomas, Hashash, Muralidhar, Saad, and Ramakrishnan]{xu2024large}
Shengzhe Xu, Christo Kurisummoottil~Thomas, Omar Hashash, Nikhil Muralidhar, Walid Saad, and Naren Ramakrishnan.
\newblock Large multi-modal models (lmms) as universal foundation models for ai-native wireless systems.
\newblock \emph{Netwrk. Mag. of Global Internetwkg.}, 38\penalty0 (5):\penalty0 10–20, July 2024.
\newblock ISSN 0890-8044.
\newblock \doi{10.1109/MNET.2024.3427313}.

\bibitem[Yu et~al.(2023)Yu, Chen, Ling, Dong, Liu, and Lu]{yu2023temporal}
Xinli Yu, Zheng Chen, Yuan Ling, Shujing Dong, Zongyi Liu, and Yanbin Lu.
\newblock Temporal data meets llm--explainable financial time series forecasting.
\newblock \emph{arXiv preprint arXiv:2306.11025}, 2023.

\end{thebibliography}
\bibliographystyle{plainnat}

\newpage
\appendix
%\onecolumn
\section{Proof of Theorem~\ref{th1}}\label{TH1}
%\begin{theorem}
\textbf{Theorem 1.} \textit{Let the logits of the ground-truth model be bounded. Then for any $f^*(k,l)$, there exists a set of functions $\{\hat{z}_i(k,l)\}_{i=1}^{|\mathcal{V}|}$ such that for all $k$ and $T_{l+1}$, the predictive distribution of the student model $\hat{p}\,(k_{T_{l+1}}\, | \, \bold{k}_{1:T_l})$ matches that of ground-truth model $p^*(k_{T_{l+1}}\,|\,\bold{k}_{1:T_l})$ and $\hat{f}(k,l) = 0$. In other words, there exists a student model with the same pre-training loss as the ground-truth model, but its logits are ineffective for the numerical downstream tasks.}
%\end{theorem}
\begin{proof}
We select $\tau \in \mathbb{R}$ such that $\forall k,T_{l+1}$, $\tau < \min_{j \in \mathcal{V}} b_j^* - \max_{j \in \mathcal{V}} z_i^*(k,l)$, and $\forall k,T_{l+1}, \forall j \in \mathcal{V}$. By setting $\hat{z}_j(k,l) := z_i^*(k,l) + \tau$, we get $\forall j \in \mathcal{V},$
\begin{align*}
 \hat{z}_j(k,l) - b_j^* < z_i^*(k,l) + \min_{j \in \mathcal{V}} b_j^* - \max_{j \in \mathcal{V}} z_i^*(k,T_{l+1}) - b_j^* \leq 0,
\end{align*}
this implies that $\sigma(\hat{z}_j(k,l) - b_j^*) = 0$. Hence, $\forall k,T_{l+1}$ and we have $\hat{f}(k,l) = 0$.
\end{proof}

\section{Proof of Lemma~\ref{lem1}}\label{lem2}
%\begin{lemma}
\textbf{Lemma 1.} \textit{Consider the Jacobian matrix $\bold{J} = \left[\frac{\partial g_i(\bold{\Psi})}{\partial \bold{\psi}_j}\right]_{i,j=1}^{|\mathcal{V}|}$, which represents the gradient of the self-attention mapping $G(\bold{\Psi})$ with respect to the input time series token embeddings. Then the spectral norm of $\bold{J}$ satisfies  
$\|\bold{J}\|_2 \leq |\bold{\Lambda}|_2 \sum_{i=1}^{|\mathcal{V}|} \left(p_{i,i} + \frac{1}{2}\right) \left|\bold{\psi}_i - \sum_{j=1}^{|\mathcal{V}|} p_{i,j} \bold{\psi}_j\right|^2 + \Delta$,  
where the residual term $\Delta$ is given by  
$\Delta = |\bold{\Lambda}|_2 \sum_{i \neq j}^{|\mathcal{V}|} p_{i,j} \left|\bold{\psi}_j - \sum_{q=1}^{|\mathcal{V}|} p_{i,q} \bold{\psi}_q\right|^2 + \frac{|\bold{\Lambda}|_2}{2} \sum_{j=1}^{|\mathcal{V}|} |\bold{\psi}_j|^2$,
and the attention weights $p_{i,j}$ are defined as  
$p_{i,j} = \frac{\exp(\bold{\psi}_i^\top \bold{\Lambda} \bold{\psi}_j)}{\sum_{k=1}^{|\mathcal{V}|} \exp(\bold{\psi}_i^\top \bold{\Lambda} \bold{\psi}_k)}$.}
%\end{lemma}
\begin{proof}
According to the analysis, the gradient of $g_i(\boldsymbol{\Psi})$ with respect to the variable $\boldsymbol{\psi}_j$ is expressed as $J_{i,j} = \frac{\partial g_i(\bold{\Psi})}{\partial \bold{\psi}_j}= p_{i,j} I + \bold{\Psi}^\top Q^i \left(\bold{\Psi} \bold{\Lambda} \delta_{i,j} + E_{j,i} \bold{\Psi} \bold{\Lambda}^\top\right)$ 
where the matrix $Q^i$ is defined by $Q^i = \text{diag}(p_{i,:}) - p_{i,:} p_{i,:}^\top$. Here, $p_{i,:} \in \mathbb{R}_+^{|\mathcal{V}|}$ corresponds to the $i$-th row of the probability matrix $\boldsymbol{P}$, $E_{j,i} \in \mathbb{R}^{|\mathcal{V}| \times |\mathcal{V}|}$ denotes a matrix with a single entry at the $(j,i)$-th position and zeros elsewhere, and $\delta_{i,j} \in \{0,1\}$ is the Kronecker delta. We thus have  
\begin{align*}
    \|\bold{J}\|_2 &\leq \sum_{i,j=1}^{|\mathcal{V}|} |J_{i,j}|_2  
\\&\leq \sum_{i,j=1}^{|\mathcal{V}|} p_{i,j} + \sum_{i=1}^{|\mathcal{V}|} |\bold{\Psi}^\top Q^i \bold{\Psi}|_2 |\bold{\Lambda}|_2 + \sum_{i,j=1}^{|\mathcal{V}|} |\bold{\Psi}^\top Q^i E_{j,i} \bold{\Psi}|_2 |\bold{\Lambda}|_2  
\\&\leq |\mathcal{V}| + |\bold{\Lambda}|_2 \sum_{i=1}^{|\mathcal{V}|} \left(\sum_{j=1}^{|\mathcal{V}|} p_{i,j} |\bold{\psi}_j|^2 - \left|\sum_{j=1}^{|\mathcal{V}|} p_{i,j} \bold{\psi}_j\right|^2\right) + |\bold{\Lambda}|_2 \sum_{i,j=1}^{|\mathcal{V}|} |\bold{\Psi}^\top Q^i e_j \bold{\psi}_i^\top|  
\\&\leq |\mathcal{V}| + |\bold{\Lambda}|_2 \sum_{i=1}^{|\mathcal{V}|} \sum_{j=1}^{|\mathcal{V}|} p_{i,j} |\bold{\psi}_j - \sum_{q=1}^{|\mathcal{V}|} p_{i,q} \bold{\psi}_q|^2 + |\bold{\Lambda}|_2 \sum_{i,j=1}^{|\mathcal{V}|} p_{i,j} |\bold{\psi}_i^\top(\bold{\psi}_j - \bold{\Psi}^\top p_{i,:})|  
\\&\leq |\bold{\Lambda}|_2 \sum_{i=1}^{|\mathcal{V}|} \left(p_{i,i} + \frac{1}{2}\right) |\bold{\psi}_i - \bold{\Psi}^\top p_{i,:}|^2 + |\mathcal{V}|  + |\bold{\Lambda}|_2 \sum_{i \neq j}^{|\mathcal{V}|} p_{i,j} |\bold{\psi}_j - \bold{\Psi}^\top p_{i,:}|^2 + \frac{|\bold{\Lambda}|_2}{2} \sum_{j=1}^{|\mathcal{V}|} |\bold{\psi}_i|^2  
\\&= |\bold{\Lambda}|_2 \sum_{i=1}^{|\mathcal{V}|} \left(p_{i,i} + \frac{1}{2}\right) |\bold{\psi}_i - \bold{\Psi}^\top p_{i,:}|^2 + |\mathcal{V}| + |\bold{\Lambda}|_2 \sum_{i \neq j}^{|\mathcal{V}|} p_{i,j} \left|\bold{\psi}_j - \sum_{q=1}^{|\mathcal{V}|} p_{i,q} \bold{\psi}_q\right|^2 + \frac{|\bold{\Lambda}|_2}{2} \sum_{j=1}^{|\mathcal{V}|} |\bold{\psi}_i|^2
%\\&= |\bold{\Lambda}|_2 \sum_{i=1}^{|\mathcal{V}|} \left(p_{i,i} + \frac{1}{2}\right) |x_i - X^\top p_{i,:}|^2 + |\mathcal{V}| + \Delta, 
\end{align*}
%where  $\Delta = |\bold{\Lambda}|_2 \sum_{i \neq j}^{|\mathcal{V}|} p_{i,j} \left|\bold{\psi}_j - \sum_{q=1}^{|\mathcal{V}|} p_{i,q} \bold{\psi}_q\right|^2 + \frac{|\bold{\Lambda}|_2}{2} \sum_{j=1}^{|\mathcal{V}|} |\bold{\psi}_i|^2.$  
%\hfill $\blacksquare$
\end{proof}
  
Theorem~\ref{th2} shows that $\bold{\Lambda}$ minimizing the objective $\sum_{i=1}^{|\mathcal{V}|} |\bold{\psi}_i - \bold{\Psi}^\top\bold{\Psi}\bold{\Lambda} \bold{\psi}_i|^2$ contains the largest $m$ eigenvectors of the correlation matrix $\bold{\Psi}^\top \bold{\Psi}$ of input time series token embeddings  where $m$ is the rank of $\bold{\Lambda}$.

Lemma 1 implies that one of the key components in the Jacobian’s upper bound takes the form  
$|\bold{\psi}_i - \sum_{j=1}^{|\mathcal{V}|} p_{i,j} \bold{\psi}_j|^2$.  
Consequently, during optimization, it is natural to aim for a reduction in the gradient magnitude, which motivates minimizing the expression  
$\sum_{i=1}^{|\mathcal{V}|} |\bold{\psi}_i - \sum_{j=1}^{|\mathcal{V}|} p_{i,j} \bold{\psi}_j|^2$.  
This leads to understand the choice of $\bold{W}^Q$ and $\bold{W}^K$ that minimize  
$\sum_{i=1}^{|\mathcal{V}|} |\bold{\psi}_i - \sum_{j=1}^{|\mathcal{V}|} p_{i,j} \bold{\psi}_j|^2$,  
which is equivalent to solving the optimization problem  
$\min_{|\bold{\Lambda}|_F \leq \rho} \sum_{i=1}^{|\mathcal{V}|} |\bold{\psi}_i - \sum_{j=1}^{|\mathcal{V}|} p_{i,j} \bold{\psi}_j|^2$,  
where the scalar constraint $\rho$ regulates the size of $\bold{\Lambda}$.

To proceed, we consider the objective in the scenario where $\rho$ is small. In this case, we can approximate the attention weights by  
$p_{i,j} \approx \frac{1}{|\mathcal{V}|} + \frac{1}{|\mathcal{V}|} \bold{\psi}_i^\top \bold{\Lambda} \bold{\psi}_j$.  
Now, we define the average of embedding as $\bar{\bold{\psi}} = \bold{\Psi}^\top 1 / |\mathcal{V}|$. It then follows that  
$\sum_{i=1}^{|\mathcal{V}|} |\bold{\psi}_i - \bold{\Psi}^\top p_{i,:}|^2 = \sum_{i=1}^{|\mathcal{V}|} |\bold{\psi}_i - \bar{\bold{\psi}} - \bold{\Psi}^\top \bold{\Psi}\bold{\Lambda} \bold{\psi}_i|^2$.  
Assuming all input time series patterns are zero-centered, i.e., $\bar{\bold{\psi}} = 0$, we have 
$\sum_{i=1}^{|\mathcal{V}|} |\bold{\psi}_i - \bold{\Psi}^\top\bold{\Psi} \bold{\Lambda} \bold{\psi}_i|^2 = \text{tr}\left((I - \bold{\Psi}^\top \bold{\Psi} \bold{\Lambda})^2 \bold{\Psi}^\top \bold{\Psi}\right)$. Theorem~\ref{th2} establishes that the optimal $\bold{\Lambda}$ that minimizes  
$\sum_{i=1}^{|\mathcal{V}|} |\bold{\psi}_i - \bold{\Psi}^\top\bold{\Psi}\bold{\Lambda} \bold{\psi}_i|^2$  
is spanned by the top $m$ eigenvectors of $\bold{\Psi}^\top \bold{\Psi}$, where $m$ equals the rank of $\bold{\Lambda}$.
\section{Proof of Theorem~\ref{th2}}\label{apth2}
%\begin{theorem}  
\textbf{Theorem 2.} \textit{Let the eigenvalues of the correlation matrix $\bold{\Psi}^\top \bold{\Psi}$ be ordered as $\lambda_1 \geq \lambda_2 \geq \cdots \geq \lambda_D$, and let $\gamma_i \in \mathbb{R}^D$ for $i = 1, \ldots, D$ denote their associated eigenvectors. Then, the matrix $\bold{\Lambda}^*$ that minimizes the quantity $\sum_{i=1}^{|\mathcal{V}|} \left|\bold{\psi}_i - \bold{\Psi}^\top\bold{\Psi} \bold{\Lambda} \bold{\psi}_i\right|^2$  
has the optimal form  
$\bold{\Lambda} = \sum_{i=1}^m \frac{1}{\lambda_i} \gamma_i \gamma_i^\top.$}
%\end{theorem}
\begin{proof}
Given that $\bold{W}_Q \in \mathbb{R}^{D \times m}$ and $\bold{W}_K \in \mathbb{R}^{D \times m}$, it follows that the matrix $\bold{\Lambda}$ has rank $m$. Hence, we know 
$\min_\bold{\Lambda} \sum_{i=1}^{|\mathcal{V}|} \|\bold{\psi}_i - \bold{\Psi}^\top \bold{\Psi} \bold{\Lambda} \bold{\psi}_i\|^2 \geq \sum_{q=m+1}^{|\mathcal{V}|} \lambda_q.$
Now, if we set $\bold{\Lambda}$ to  
$\bold{\Lambda} = \sum_{i=1}^m \frac{1}{\lambda_i} \gamma_i \gamma_i^\top$,  
then we obtain  
$\sum_{i=1}^{|\mathcal{V}|} \|\bold{\psi}_i - \bold{\Psi}^\top \bold{\Psi} \bold{\Lambda} \bold{\psi}_i\|^2 = \text{tr}\left(\left(I - \sum_{i=1}^m \gamma_i \gamma_i^\top\right)^2 \bold{\Psi}^\top \bold{\Psi}\right) = \sum_{q=m+1}^D \lambda_q.$

Therefore, the optimal solution $\bold{\Lambda}$ for minimizing $\sum_{i=1}^{|\mathcal{V}|} \|\bold{\psi}_i - \bold{\Psi}^\top \bold{\Psi} \bold{\Lambda} \bold{\psi}_i\|^2$ is essentially characterized as a linear combination of the top $m$ eigenvectors of $\bold{\Psi}^\top \bold{\Psi}$. Since a small gradient will prefer a small quantity of $\sum_{i=1}^{|\mathcal{V}|} \|\bold{\psi}_i - \bold{\Psi}^\top \bold{\Psi} \bold{\Lambda} \bold{\psi}_i\|^2$, the self-attention mechanism implicitly drives the weight matrices $\bold{W}_Q$ and $\bold{W}_K$ to align with the dominant eigen-directions of $\bold{\Psi}^\top \bold{\Psi}$.
\end{proof}

\section{Clustering in the Contextual Embedding Space}\label{cluster_appendix}
\textbf{Clustering.} We begin with the isotropy assesmment by performing clustering on the LLM representations in the contextual embedding space. There are various methods for performing cultering, such as $k$-means, DBSCAN \cite{DBS}. We select $K$-means clustering method because it is reasonably fast in high embedding dimensions (e.g., $d\geq768$ for GPT2, ELMo, BERT etc.). We use the celebrated silhouette score analysis \cite{ROUSSEEUW198753} to determine the number of clusters $|C|$ in the contextual embedding space. After performing $K$-means clustering, each observation $p$ (i.e., one of the $\bold{J}$ vector representations in $\mathcal{V}$) is assigned to one of $C$ clusters. For an observation $p$ assigned to the cluster $c\in C$, we compute the silhouette score as follows
\begin{align*}
    a(p) = \frac{1}{\left|C\right|-1}\sum \limits_{q \in C , p \neq q}\textrm{dist}(p,q); \qquad
    b(p) = \min_{ \Tilde{c}\neq c}\sum \limits_{q\in \Tilde{c}}\textrm{dist}(p,q); \qquad
    s(p) = \frac{b(p)-a(p)}{\max b(p),a(p)},
\end{align*}
where $a(p)$ is the mean distance between an observation $p$ and the rest in the same cluster class $p$, while $b(p)$ measures the smallest mean distance from $p$-th observation to all observations in the other cluster class. After computing the silhouette scores $s(p)$ of all observations, a global score is computed by averaging the individual silhouette values, and the partition (with a specific number of clusters) of the largest average score is pronounced superior to other partitions with a different number of clusters. We select the best $|C|$ that belongs to the partition that scores highest among the other partitions.

\section{Dataset Description}\label{data_des}
\vspace{-0.05in}
\textbf{Real Datsets.}
\vspace{-0.1in}
\begin{table}[htb!]
\centering
\caption{The complete list of datasets used for our quantitative and qualitative analysis. The table is divided
into three sections, representing how the datasets were used for baseline models.}
\label{tab:all-datasets}
\scalebox{0.87}{
\begin{tabular}{lllrrrrr}
\toprule
\multirow{2}{*}{\textbf{Dataset}} & \multirow{2}{*}{\textbf{Domain}} & \multirow{2}{*}{\textbf{Freq.}} & \multirow{2}{*}{\textbf{Num. Series}} &    \multicolumn{3}{c}{\textbf{Series Length}}     & \multicolumn{1}{c}{\textbf{Prediction}}\\
\cmidrule(lr){5-7}
 &        &           &            &      \textbf{min}     &  \textbf{avg}     &     \textbf{max}     & \multicolumn{1}{c}{\textbf{Length ($H$)}} \\
\midrule
\href{https://zenodo.org/record/4659727}{Australian Electricity} & Energy & 30min & 5 & 230736 & 231052 & 232272 & 48 \\
\href{https://zenodo.org/record/4656022}{Car Parts} & Retail & 1M & 2674 & 51 & 51 & 51 & 12 \\
\href{https://zenodo.org/record/4656009}{Covid Deaths} & Healthcare & 1D & 266 & 212 & 212 & 212 & 30 \\
\href{https://www.chicagobooth.edu/research/kilts/research-data/dominicks}{Dominick} & Retail & 1D & 100014 & 201 & 296 & 399 & 8 \\
\href{https://github.com/laiguokun/multivariate-time-series-data/tree/master/exchange_rate}{Exchange Rate} & Finance & 1B & 8 & 7588 & 7588 & 7588 & 30 \\
\href{https://zenodo.org/records/4654833}{FRED-MD} & Economics & 1M & 107 & 728 & 728 & 728 & 12 \\
\href{https://zenodo.org/record/4656014}{Hospital} & Healthcare & 1M & 767 & 84 & 84 & 84 & 12 \\
\href{https://zenodo.org/records/4656125}{NN5 (Weekly)} & Finance & 1W & 111 & 113 & 113 & 113 & 8 \\
\href{https://zenodo.org/record/4654822}{Weather} & Nature & 1D & 3010 & 1332 & 14296 & 65981 & 30 \\
\href{https://zenodo.org/record/4654822}{Transportaion Signal} & Transport & 1D & 3010 & 1332 & 14296 & 65981 & 30 \\
\midrule
Synthetic (10 kernels) & Numerical & - &1000000& 1024 & 1024&1024& 64\\
\bottomrule
\end{tabular}
}
\end{table}

\textbf{Synthetic Datasets.}
We use KernelSynth~\cite{ansari2024chronos}, a method to generate synthetic dataset using Gaussian processes (GPs). KernelSynth allows generation of large, diverse datasets tailored to specific patterns or statistical properties, which is particularly useful when real-world data is scarce or incomplete. In this synthetic data generation process, the GPs are defined by a mean function, $\mu(t)$, and a positive definite kernel, $\kappa(x_i, x_j)$, which specifies a covariance function for variability across input pairs $(x_i, x_j)$. A kernel bank $\mathcal{K}$ (which consists of linear, RBF, and periodic kernels) is used to define diverse time series patterns. The final kernel $\tilde{\kappa}(x_i, x_j)$ is constructed by sampling and combining kernels from $\mathcal{K}$ using binary operations like $+$ and $\times$. Synthetic time series are generated by sampling from the GP prior, $GP(\mu(t) = 0, \tilde{\kappa}(x_i, x_j))$.
%\begin{comment}
The following algorithm presents the pseudocode for KernelSynth which essentially follows the approach in~\cite{ansari2024chronos}.
\begin{algorithm}[H]\label{alg1}
\caption{\textsc{KernelSynth}: Generating Synthetic Sequences via Gaussian Process Kernels}
\textbf{Input}: Kernel bank $\mathcal{K}$, maximum kernels per time series $J = 5$, and length of the time series $l_{\text{syn}} = 1024$. \\
\textbf{Output}: A synthetic time series $\mathbf{x}_{1:l_{\text{syn}}}$.
\begin{algorithmic}[1]
\State $j \sim \mathcal{U}\{1, J\}$ \Comment{sample the number of kernels}
\State $\{\kappa_1(t, t'), \ldots, \kappa_j(t, t')\} \overset{\text{i.i.d}}{\sim} \mathcal{K}$ \Comment{sample $j$ kernels from the Kernel bank $\mathcal{K}$}
\State $\kappa^*(t, t') \leftarrow \kappa_1(t, t')$
\For{$i \leftarrow 2$ to $j$}
    \State $\star \sim \{+, \times\}$ \Comment{pick a random operator (add or multiply)}
    \State $\kappa^*(t, t') \leftarrow \kappa^*(t, t') \star \kappa_i(t, t')$ \Comment{compose kernels}
\EndFor
\State $\mathbf{x}_{1:l_{\text{syn}}} \sim \mathcal{GP}(0, \kappa^*(t, t'))$ \Comment{draw a sample from the GP prior}
\State \Return{$\mathbf{x}_{1:l_{\text{syn}}}$}
\end{algorithmic}
\end{algorithm}
%\end{comment}

\section{Full Visualization of PCA plots for different models}\label{full_pca_plots}
\vspace{-0.05in}
%The Variations in Chornos-T5's hidden representations for different input context lengths is depicted in Figure~\ref{pca_plot_context}
%}
%\afterpage{
\begin{comment}
\begin{figure}
\centering
\includegraphics[width=0.8\textwidth]{figures/RBF_PCA.pdf}
%\vspace{-10pt}
\caption{Variations in Chronos-T5's hidden representations for different input context lengths for the same synthetic dataset ``non-linear-Dataset 1''
 : (a) Contextual embedding space for input context length $ L = 500$. (b) Contextual embedding space for input context length $L = 100$. 
}
\label{pca_plot_context}
\end{figure}
\end{comment}
%\clearpage
The full visualization of PCA plots of different models is provided below. We use the synthetic  Dataset 1, and Dataset 2 from non-linear domain for illustration. 
\begin{figure*}[ht]
\centering
\includegraphics[width=1\linewidth]
{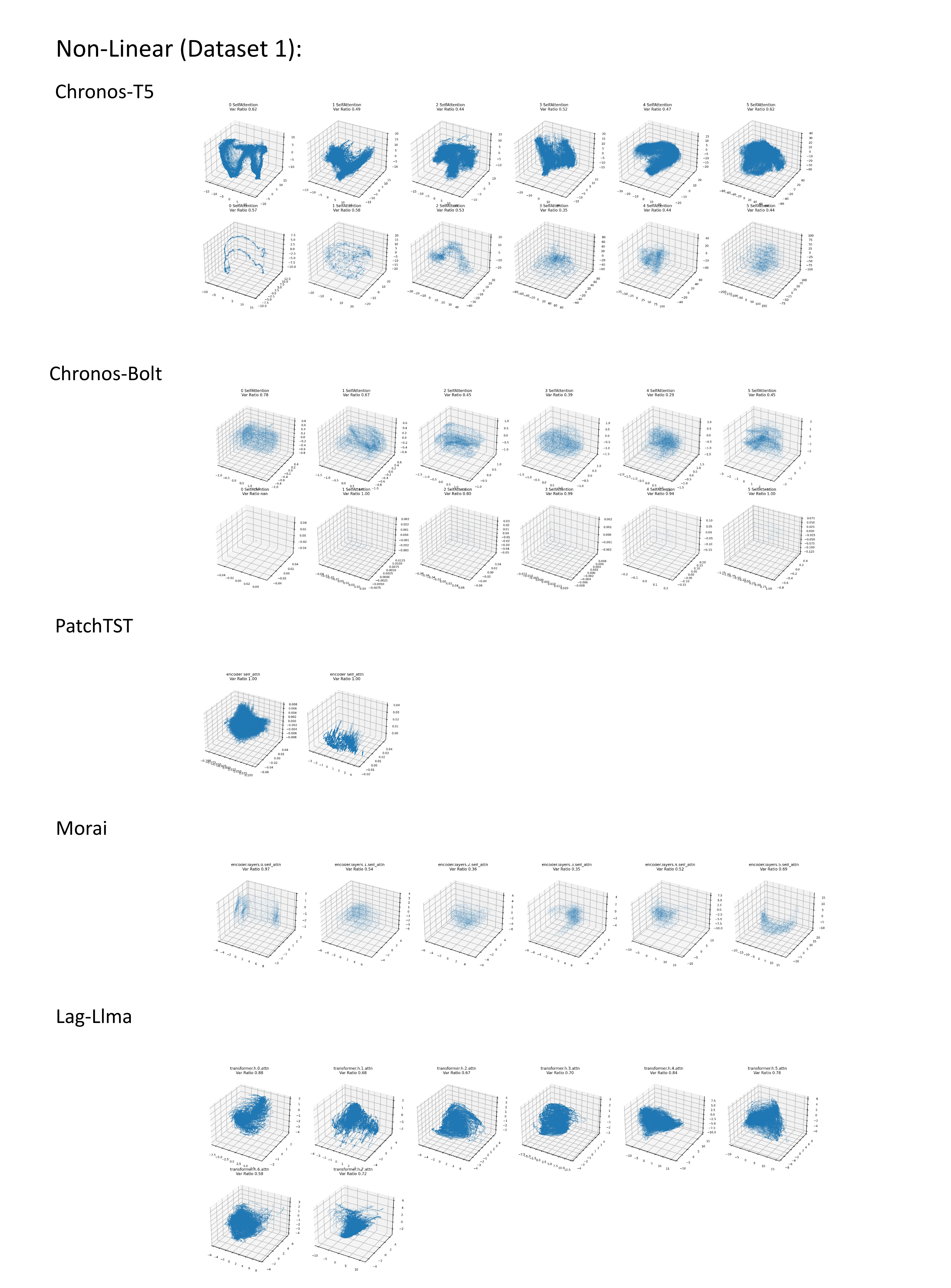}
%\label{cluster_fig}
\vspace{-0.14in}
\end{figure*}
%}
\begin{figure*}[ht]
\centering
\includegraphics[width=1.0\linewidth]
{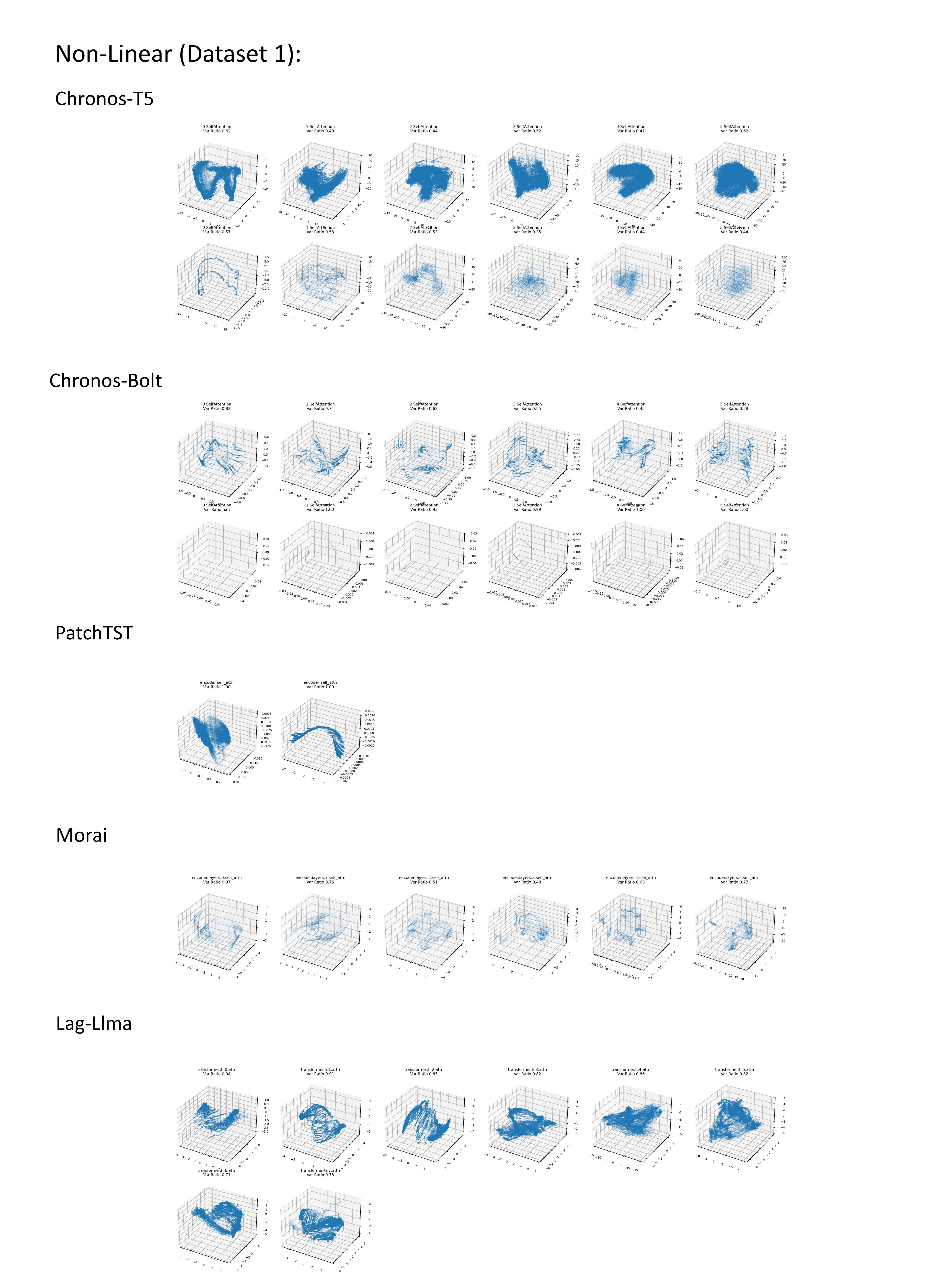}
%\label{cluster_fig}
\vspace{-0.14in}
\end{figure*}
%}

\end{document}